\documentclass{article} 
\usepackage{iclr2027_conference,times}

\usepackage{amsmath,amsfonts,bm}

\def\eqref#1{equation~\ref{#1}}
\def\Eqref#1{Equation~\ref{#1}}

\def\1{\bm{1}}

\def\rvepsilon{{\mathbf{\epsilon}}}

\def\rvx{{\mathbf{x}}}

\def\vzero{{\bm{0}}}

\def\vmu{{\bm{\mu}}}
\def\vtheta{{\bm{\theta}}}

\def\vv{{\bm{v}}}

\def\vx{{\bm{x}}}

\def\evlambda{{\lambda}}

\def\evmu{{\mu}}

\def\evv{{v}}

\def\mA{{\bm{A}}}

\def\mD{{\bm{D}}}

\def\mI{{\bm{I}}}

\def\mL{{\bm{L}}}

\def\mU{{\bm{U}}}

\def\mW{{\bm{W}}}

\def\mSigma{{\bm{\Sigma}}}

\DeclareMathAlphabet{\mathsfit}{\encodingdefault}{\sfdefault}{m}{sl}
\SetMathAlphabet{\mathsfit}{bold}{\encodingdefault}{\sfdefault}{bx}{n}

\def\gG{{\mathcal{G}}}

\newcommand{\E}{\mathbb{E}}

\newcommand{\Var}{\mathrm{Var}}

\renewcommand{\rvepsilon}{\bm{\epsilon}}
\usepackage{amsthm}
\newtheorem{proposition}{Proposition}
\newtheorem{corollary}{Corollary}
\usepackage{hyperref}
\usepackage{url}
\usepackage{graphicx}

\title{Graph Residual Conjugate Diffusion:\newline SNR-Equalized Heat Flow for Graph Signals}

\author{Jinwei Li \\
Department of Data Science\\
Friedrich-Alexander-Universität Erlangen-Nürnberg\\
Erlangen, Germany\\
\texttt{jinwei.li@fau.de}
\And
Daniel Tenbrinck \\
Department of Data Science\\
Friedrich-Alexander-Universität Erlangen-Nürnberg\\
Erlangen, Germany\\
\texttt{daniel.tenbrinck@fau.de}
}

\iclrfinalcopy 
\begin{document}

\maketitle




\begin{abstract}
Diffusion models generate data by reversing a forward corruption process that typically approaches a simple Gaussian prior. 
Recent work has extended this framework to signals supported on fixed graphs, e.g., road-network traffic and sensor-network measurements.
Many graph signals have nonuniform spectral energy across graph frequencies, whereas isotropic corruption adds the same conditional noise variance to every graph-frequency mode. 
Driving all modes to near-zero terminal signal-to-noise ratio (SNR) requires strong corruption, which increases the noise range that must be covered under a fixed sampling budget.
We introduce \emph{Graph Residual Conjugate Diffusion} (GRCD), which replaces the shared clock of graph heat diffusion with a mode-dependent clock that gives every graph-Fourier mode the same conditional signal-to-noise ratio.
GRCD fits a zero-mean graph-spectral Gaussian reference on the training split and, rather than driving the corruption to near-zero terminal SNR, stops at a finite terminal SNR at which the propagated reference still carries the fitted spectral variances, narrowing the SNR range that sampling must cover.
The Gaussian component has an exact modewise propagator in the probability-flow ODE, so sampling advances it analytically and integrates only the learned residual score numerically.
We evaluate GRCD on five graph-signal settings (the METR-LA traffic and Molene weather datasets, and three synthetic settings from stochastic block models) against seven comparators under a matched protocol: Graph-Aware Diffusion (GAD), EDM \citep{karras2022edm} adapted to the graph backbone, two graph adaptations of Whitened Score Diffusion (WSD), and three preconditioning controls.
At four function evaluations (NFEs), GRCD lowers averaged maximum mean discrepancy (aMMD) by $22\text{--}36\times$ over the best comparator on all five settings, reaching $0.054$ on METR-LA, where it clears an aMMD $0.1$ target with $87\%$ less sampling wall-clock time than the cheapest comparator that reaches it. Within GRCD, fitting the terminal reference reduces aMMD by factors of $2.7\text{--}7.3$ at finite terminal SNR, while the corresponding factors shrink to $1.00\text{--}1.01$ near zero, matching the analytic limit in which the propagated reference loses dependence on the fitted covariance.
\end{abstract}

\section{Introduction}

Diffusion models combine a prescribed forward corruption process with learned reverse dynamics, while the forward process is chosen so that its terminal marginal approaches a simple tractable reference distribution, most commonly a Gaussian distribution \citep{ho2020ddpm,song2021sde}.

Graph signals can have strongly nonuniform spectral energy, especially when they are smooth with respect to the underlying graph topology \citep{ortega2018graph,rozada2026gad}.
In our benchmarks this appears as a large variation in graph-Fourier coefficient variance across frequencies.
However, isotropic corruption injects the same noise variance into every graph-Fourier mode \citep{ho2020ddpm,song2021sde} independently of the spectral graph characteristics.
Graph-Aware Diffusion (GAD) \citep{rozada2026gad} incorporates the graph structure through graph heat diffusion, while Whitened Score Diffusion (WSD) \citep{alido2025wsd} uses anisotropic Gaussian corruption and structured spectral priors. 
In this work, we study a complementary question: whether a graph-signal diffusion process should be driven to an endpoint near zero SNR when a structured finite-SNR reference can instead be fitted and propagated analytically.

We derive a mode-dependent clock, i.e., a frequency-dependent diffusion rate, for which the time-dependent graph-filter transformation
\(\widetilde{\rvx}_t=\mA_t^{-1}\rvx_t\) reduces the forward process to isotropic Gaussian corruption. 
In these conjugate coordinates every graph-Fourier mode has unit signal coefficient and conditional noise variance \(q(t)^2\), hence the same conditional SNR \(q(t)^{-2}\).
Thus, a single scalar \(q(t)\) controls the spectral corruption level. 
We call this construction a \emph{conjugate forward process}.

PriorGrad showed that data-dependent Gaussian priors can improve conditional diffusion \citep{lee2022priorgrad}. 
Instead of a prior that varies with the conditioning input, we fit one zero-mean Gaussian reference shared by every sample: we estimate a Ledoit--Wolf shrinkage covariance \citep{ledoit2004covariance}, retain its diagonal variances in the graph-Fourier basis, and propagate the resulting reference to the chosen finite terminal SNR.
This defines the initial distribution for reverse sampling.

We use the probability-flow ODE \citep{song2021sde} and decompose its score
into an analytic Gaussian reference component and a learned residual.
Related decompositions are known: \citet{remy2023probabilistic} combine an analytic Gaussian score with a learned residual trained by residual denoising score matching, while \citet{wang2024gaussian} show that learned diffusion scores are well approximated by a Gaussian linear score at moderate to high noise levels and exploit the corresponding analytic dynamics for sampling acceleration. 
In our method, the Gaussian reference is fitted in the graph-Fourier basis and propagated under conjugate graph diffusion, so its score is available in closed form at every time.
This reference also yields our sampler's exact modewise propagator.

\paragraph{Contributions:} We propose \emph{Graph Residual Conjugate Diffusion} (GRCD), decomposing the score into a Gaussian reference score and a learned residual correction.
In graph-Fourier coordinates the Gaussian contribution has a closed-form modewise propagator, so only the residual is integrated numerically.
The forward process, the fitted reference, the residual parameterization, and the exponential-residual solver are separable components of the diffusion process, allowing their effects to be evaluated through controlled ablations.
Our empirical evaluation focuses on low-budget graph-signal generation and evaluates these components individually.
Our main contributions are:

\textbf{Conjugate graph diffusion.}
We derive a mode-dependent clock whose conjugate coordinates have equal conditional noise variance across graph frequencies, providing a single scalar corruption coordinate while retaining graph-dependent dynamics in the original signal space.\\
\textbf{Fitted finite-SNR reference and residual propagation.}
We fit a Gaussian reference in the graph-Fourier basis to training data only and propagate it analytically to finite terminal SNR for reverse initialization.
The same reference yields a closed-form propagator for the Gaussian component of the probability-flow ODE, leaving only the learned residual to be integrated numerically.\\
\textbf{Asymptotic behavior at zero terminal SNR.}
We show analytically that the propagated reference becomes independent of the fitted spectral covariance as the terminal noise grows, converging to \(\mathcal{N}(\vzero,\sigma^2\mL_\delta^{-1})\), where \(\mL_\delta\) is the shifted, spectrally normalized graph Laplacian from Section~\ref{sec:background}. The experiments match this prediction: across five settings, fitting the terminal reference reduces aMMD by factors of $2.7$--$7.3$ at finite terminal SNR, while the factors shrink to $1.00$--$1.01$ near zero.\\
\textbf{Low-budget generation.}
On METR-LA, GRCD reaches an aMMD of $0.054$ with only $4$ function evaluations (NFE), which is lower than every external baseline and preconditioning control at evaluated NFE budgets from $4$ to $64$.
For an aMMD target of $0.10$, GRCD reduces sampling wall-clock time by $87\%$ relative to the cheapest of these that reaches the target.

\section{Background and setting}
\label{sec:background}
\paragraph{Graph signals.}
We consider a fixed, undirected weighted graph $\gG$ on $N \in \mathbb{N}$ nodes.
Let $\mW$ denote the symmetric weighted adjacency matrix and $\mD$ the diagonal degree matrix with \(\mD_{i,i}=\sum_{j=1}^N \mW_{i,j}\).
Following the standard graph-signal-processing construction \citep{luxburg2007tutorial,shuman2013emerging,ortega2018graph,rozada2026gad}, we use the symmetric normalized graph Laplacian
\(
\mL
=
\mI-\mD^{-1/2}\mW\mD^{-1/2}.
\)
For an undirected graph with nonnegative weights, $\mL$ is symmetric positive semidefinite \citep{luxburg2007tutorial}, and therefore admits the eigendecomposition
\(
\mL
=
\mU\operatorname{diag}(\bm{\lambda})\mU^\top,
\mU^\top\mU=\mI,
\evlambda_i\geq0.
\)
For the diffusion dynamics, we normalize the Laplacian by its largest eigenvalue \(\lambda_{\max}\) to place the graph spectrum on a common scale across datasets, and add a small positive spectral shift \(\delta>0\) so that all eigenvalues are strictly positive: 
\(
\mL_\delta
=
\frac{\mL}{\lambda_{\max}}
+
\delta\mI
=
\mU\operatorname{diag}(\vmu)\mU^\top,
\evmu_i
=
\frac{\evlambda_i}{\lambda_{\max}}
+
\delta
>0.
\)
Because $\mL_\delta$ is an affine function of $\mL$, it has the same eigenvectors and preserves the ordering of the Laplacian eigenvalues.
\paragraph{Score-based diffusion.}
Let $p_t$ denote the forward diffusion probability density at time $t$.
A score-based diffusion model learns the time-dependent score vector
\(
\nabla_{\vx}\log p_t(\vx).
\)
For the corresponding SDE, there exists a deterministic probability-flow ODE that, when initialized from the same distribution, has the same marginal density \(p_t\) at every time \(t\) \citep{song2021sde}.
We use this probability-flow formulation for sampling and measure sampling cost in NFE.

\paragraph{Terminal SNR.}
The SNR provides a natural parameterization of diffusion noise levels \citep{kingma2021variational}.
\citet{lin2024common} showed for image diffusion models that a nonzero terminal SNR can create a training--inference mismatch when inference is initialized from pure noise.
In GRCD, terminal SNR plays a different role: it controls how much fitted graph-spectral covariance remains in the propagated terminal reference.
We define the corresponding quantity in Section~\ref{sec:gaussian_reference}.

\section{Methodology}
\label{sec:method}
\subsection{Conjugate graph diffusion}
\label{sec:conjugate_forward}

Motivated by graph heat-diffusion models such as Graph-Aware Diffusion (GAD) \citep{rozada2026gad}, we seek a graph diffusion with a mode-dependent clock that is exactly conjugate to an isotropic variance-exploding (VE) process.
Throughout this section \(\rvx_t\in\mathbb{R}^N\) denotes the graph signal at diffusion time \(t\) and
\(
x_{t,i}:=[\mU^\top\rvx_t]_i
\)
denotes its \(i\)-th graph-Fourier coefficient. \(\evmu_i\) is the eigenvalue of the shifted normalized Laplacian \(\mL_\delta\) associated with this mode. 
We model each graph-Fourier coefficient with the linear stochastic differential equation (SDE)
\begin{equation}
dx_{t,i}
\: = \:
-\evmu_i c_i(t)x_{t,i}\,dt
+
\sqrt{2\sigma^2 c_i(t)}\,dW_{t,i},
\label{eq:sde}
\end{equation}
where \(W_{t,i}:=[\mU^\top\bm W_t]_i\) is a standard one-dimensional Brownian motion, independent across modes, \(c_i(t)\geq0\) is the mode-dependent clock rate defined below, and
\(\sigma>0\) is a fixed diffusion scale. The drift term
\(-\evmu_i c_i(t)x_{t,i}\,dt\) damps mode \(i\) toward zero, while the stochastic term injects Gaussian noise with
variance rate \(2\sigma^2 c_i(t)\).
The clock \(c_i(t)\) controls each spectral mode's evolution.

Let \(q(t)=\kappa t\) for \(t\in[0,t_{\max}]\) and \(\kappa>0\), with training and sampling restricted to \(t\in[t_{\min},t_{\max}]\). In the conjugate coordinates, we use the isotropic VE process
\(
d\widetilde{\rvx}_t
=
\sqrt{2q(t)q'(t)}\,d\bm{W}_t,
\)
whose conditional marginals at each fixed \(t\) admit the reparameterization
\begin{equation}
\widetilde{\rvx}_t
\: = \:
\widetilde{\rvx}_0
+
q(t)\rvepsilon,
\qquad
\rvepsilon\sim\mathcal{N}(\vzero,\mI).
\label{eq:conjugate_isotropic_main}
\end{equation}

\begin{proposition}[Graph-filter conjugate representation]
\label{prop:conjugacy}
For the isotropic VE process introduced above with \(\widetilde{\rvx}_0=\rvx_0\), define
\begin{equation}
\mA_t
\: := \:
\left(
\mI
+
\frac{q(t)^2}{\sigma^2}\mL_\delta
\right)^{-1/2},
\label{eq:At}
\end{equation}
and let
\(
\rvx_t=\mA_t\widetilde{\rvx}_t.
\)
The matrix \(\mA_t\) has the same eigenvectors as \(\mL_\delta\), with
eigenvalue
\(
a_i(t)
=
\left(
1+\frac{\evmu_i q(t)^2}{\sigma^2}
\right)^{-1/2}
\)
in graph-Fourier mode \(i\). The resulting process satisfies
\begin{equation}
x_{t,i}
\: = \:
a_i(t)
\left(
x_{0,i}
+
q(t)\epsilon_i
\right),
\qquad
\epsilon_i:=[\mU^\top\rvepsilon]_i,
\label{eq:forward}
\end{equation}
and hence
\(
\Var\!\left(
x_{t,i}
\mid
x_{0,i}
\right)
=
a_i(t)^2q(t)^2.
\)
In conjugate coordinates,
\(
\Var\!\left(
\widetilde{x}_{t,i}
\mid
\widetilde{x}_{0,i}
\right)
=
q(t)^2
\)
for every mode \(i\), so all graph-Fourier modes carry the same conditional noise variance.
\end{proposition}

\begin{corollary}[Induced mode-dependent clock]
\label{cor:modewise_sde}
The graph-Fourier modes of the process described in
Proposition~\ref{prop:conjugacy} satisfy \Eqref{eq:sde} with
\begin{equation}
c_i(t)
\: = \:
\frac{q(t)q'(t)}
{\sigma^2+\evmu_iq(t)^2}.
\label{eq:Mode-dependent_clock}
\end{equation}
\end{corollary}

The conditional noise variance induced in graph coordinates is
\(
a_i(t)^2q(t)^2
=
\frac{\sigma^2 q(t)^2}
{\sigma^2+\evmu_i q(t)^2},
\)
which approaches \(\sigma^2/\evmu_i\) as \(q(t)\to\infty\), whereas the corresponding variance \(q(t)^2\) in conjugate coordinates grows without bound.
At the same time, the signal coefficient \(a_i(t)\) tends to zero, and the conditional SNR, in which \(a_i(t)\) cancels, equals \(1/q(t)^2\) and likewise tends to zero. Thus, although the noise variance remains bounded in graph coordinates, the forward process still removes dependence on the initial signal in the large-\(q(t)\) limit. Section~\ref{sec:gaussian_reference} instead uses a finite terminal noise level so that data-dependent spectral structure can remain in the terminal reference.
The proof of Proposition~\ref{prop:conjugacy} and the derivation of Corollary~\ref{cor:modewise_sde} are provided in Appendix~\ref{app:conjugacy_proof}.

\subsection{Gaussian reference and finite-SNR terminal initialization}
\label{sec:gaussian_reference}
We fit a zero-mean Gaussian reference using only the training split. Let $\mSigma_{\mathrm{LW}}$ denote the Ledoit--Wolf shrinkage covariance \citep{ledoit2004covariance}.
We transform this covariance to the graph-Fourier basis and retain only its diagonal entries, discarding cross-mode covariances:
\begin{equation}
\evv_i
\: = \:
\left[
\mU^\top
\left(
\mSigma_{\mathrm{LW}}
+\varepsilon_{\mathrm{ref}}\mI
\right)
\mU
\right]_{ii},
\qquad
\mSigma_{\mathrm{ref}}
\: = \:
\mU\operatorname{diag}(\vv)\mU^\top.
\label{eq:ref}
\end{equation}
Here, \(\varepsilon_{\mathrm{ref}}>0\) is a small numerical shift. 
Since \(\mSigma_{\mathrm{LW}}\) is positive semidefinite and \(\mU\) is orthonormal,
\(
\evv_i
=
\left[
\mU^\top
\mSigma_{\mathrm{LW}}
\mU
\right]_{ii}
+
\varepsilon_{\mathrm{ref}}
\geq
\varepsilon_{\mathrm{ref}},
\)
so every spectral variance is strictly positive and \(\mSigma_{\mathrm{ref}}\) is non-singular. Thus, \(\mSigma_{\mathrm{ref}}\) defines the Gaussian reference distribution \(\mathcal{N}(\vzero,\mSigma_{\mathrm{ref}})\). Further details of the spectral diagonal approximation are given in Appendix~\ref{app:reference}.

Because $\mA_t$ and $\mSigma_{\mathrm{ref}}$ are diagonal in the same graph-Fourier basis, the reference propagates analytically under the conjugate forward process.
Its variance in mode $i$ is
\begin{equation}
\gamma_i(t)
\: = \:
a_i(t)^2\left(\evv_i+q(t)^2\right)
\: = \:
\frac{\sigma^2\!\left(\evv_i+q(t)^2\right)}
{\sigma^2+\evmu_iq(t)^2}.
\label{eq:gamma}
\end{equation}
Therefore, we get the exact marginal of the chosen Gaussian reference distribution
\[
p_{\mathrm{ref},t}
\: = \:
\mathcal{N}\!\left(
\vzero,\,
\mU\operatorname{diag}\!\left(
\gamma_1(t),\ldots,\gamma_N(t)
\right)
\mU^\top
\right).
\]
The forward kernel itself does not depend on the fitted spectral variances \(\evv_i\).
By Proposition~\ref{prop:conjugacy} mode \(i\) has signal coefficient
\(a_i(t)\) and conditional variance \(a_i(t)^2q(t)^2\), so
\begin{equation}
\mathrm{SNR}^{\mathrm{cond}}_i(t)
\: = \:
\frac{
a_i(t)^2
}{
a_i(t)^2 q(t)^2
}
\: = \:
\frac{1}{q(t)^2}
\label{eq:snr_conditional}
\end{equation}
is the same for every graph-Fourier mode; this is the SNR definition of
\citet{kingma2021variational} applied modewise. The coefficient \(a_i(t)\)
cancels, so the mode-dependent clock equalizes conditional corruption across
graph frequencies.

A second, reference-level signal-to-noise ratio describes the fitted reference
rather than the forward kernel: it measures how much fitted data-dependent
signal remains relative to the injected noise, and therefore depends on the
data. Under the reference, the propagated signal and injected-noise variances
in graph-Fourier mode \(i\) are \(a_i(t)^2\evv_i\) and \(a_i(t)^2q(t)^2\),
respectively, so the reference SNR in mode $i$ at time $t$ is given by
\begin{equation}
\mathrm{SNR}^{\mathrm{ref}}_i(t)
\: = \:
\frac{a_i(t)^2\evv_i}
{a_i(t)^2q(t)^2}
\: = \:
\frac{\evv_i}{q(t)^2}.
\label{eq:snr_mode_main}
\end{equation}
For reporting a single terminal value we use the energy-weighted average
\(
\overline{\mathrm{SNR}}^{\mathrm{ref}}(t)
=
\sum_{i=1}^N
\frac{\evv_i}{\sum_{j=1}^N \evv_j}\,
\mathrm{SNR}^{\mathrm{ref}}_i(t),
\)
so that modes carrying little signal energy cannot dominate the summary.
Therefore, as the terminal noise level \(q(t_{\max})=\kappa t_{\max}\) grows, \(\mathrm{SNR}^{\mathrm{ref}}_i(t_{\max})\to0\), and we consequently have
\begin{equation}
\gamma_i(t_{\max})
\: = \:
\sigma^2
\frac{
\evv_i/q(t_{\max})^2+1
}{
\sigma^2/q(t_{\max})^2+\evmu_i
}
\quad \xrightarrow[]{q(t_{\max})\to\infty}\quad 
\frac{\sigma^2}{\evmu_i}.
\label{eq:gmrf_limit_main}
\end{equation}
Thus, since \(\evmu_i>0\) holds for every mode, dependence on the fitted spectral variance \(\evv_i\) vanishes in the limit, leaving the limiting variance \(\sigma^2/\evmu_i\).

\subsection{Residual score parameterization}
\label{sec:residual_score}

The propagated Gaussian reference distribution has the following graph-Fourier score
\[
s_{\mathrm{ref},i}(\vx,t)
\: = \:
-\frac{x_i}{\gamma_i(t)},
\qquad
x_i \: = \: [\mU^\top\vx]_i.
\]
Let
\(
s_i^*(\vx,t)
\: := \:
\left[
\mU^\top\nabla_{\vx}\log p_t(\vx)
\right]_i
\)
denote the unknown marginal data score in mode \(i\). We define the residual $r_i^*$ as
\(
r_i^*(\vx,t)
:=
s_i^*(\vx,t)-s_{\mathrm{ref},i}(\vx,t).
\)
We parameterize this residual with a neural network while keeping the Gaussian reference score in closed form.

Let \(\eta_i(t)=q(t)a_i(t)\) denote the conditional standard deviation in mode \(i\). Using the definition of \(a_i(t)\), we get
\(
\eta_i(t)
=
q(t)
\left(
1+\frac{\evmu_i q(t)^2}{\sigma^2}
\right)^{-1/2}
=
\frac{\sigma q(t)}
{\sqrt{\sigma^2+\evmu_i q(t)^2}}.
\)
For \(q(t)\geq0\), \(\eta_i(t)\) increases with \(q(t)\) and approaches the finite positive limit
\(
\eta_i(t)
\longrightarrow
\frac{\sigma}{\sqrt{\evmu_i}}
\ \text{as }q(t)\to\infty,
\)
where \(\evmu_i>0\) is ensured by the shift \(\delta>0\). Thus, the residual scaling \(1/\eta_i(t)\) remains finite in the large-noise limit.

Let \(f_{\vtheta}(\vx,t)\) denote the network output in the node basis and
\(
f_{\vtheta,i}(\vx,t)
\: = \:
[\mU^\top f_{\vtheta}(\vx,t)]_i
\)
its \(i\)-th graph-Fourier coefficient.
We parameterize the residual as
\(
r_{\vtheta,i}(\vx,t)
\: = \:
-\frac{f_{\vtheta,i}(\vx,t)}{\eta_i(t)}.
\)
The reconstructed score is then given as
\begin{equation}
s_{\vtheta,i}(\vx,t)
\: = \:
-\frac{x_i}{\gamma_i(t)}
-
\frac{f_{\vtheta,i}(\vx,t)}{\eta_i(t)}.
\label{eq:score}
\end{equation}

For the forward corruption of graph signals induced by the conjugate construction, the corresponding noise-prediction target is
\(
f_i^*
\: = \:
\epsilon_i
-
\eta_i(t)
\frac{x_{t,i}}{\gamma_i(t)}.
\)
We train with the mean-squared denoising objective
\citep{vincent2011connection,ho2020ddpm}
\begin{equation}
\mathcal{L}(\vtheta)
\: = \:
\E_{t,\rvx_0,\rvepsilon}
\left[
\frac{1}{N}
\sum_{i=1}^N
\left(
f_{\vtheta,i}(\rvx_t,t)-f_i^*
\right)^2
\right],
\label{eq:loss}
\end{equation}
where \(t\sim\mathcal{U}[t_{\min},t_{\max}]\). Appendix~\ref{app:residual_target} shows that the population minimizer of \Eqref{eq:loss}, together with \Eqref{eq:score}, recovers the marginal score \(s_i^*(\vx,t)\) exactly.
Thus, the residual parameterization changes the denoising target's representation rather than approximating the marginal score.

\subsection{Exact Gaussian propagation and the exponential-residual sampler}
\label{sec:exp_residual_sampler}

For the modewise SDE in \Eqref{eq:sde}, define
\(b_i(t):=-\evmu_i c_i(t)\) and
\(g_i(t)^2:=2\sigma^2c_i(t)\).
Substituting \Eqref{eq:score} into the probability-flow ODE
\citep{song2021sde} gives
\begin{equation}
\frac{dx_i}{dt}
\: = \:
\underbrace{
\left[
b_i(t)+\frac{g_i(t)^2}{2\gamma_i(t)}
\right]x_i
}_{\text{Gaussian reference}}
-
\underbrace{
\frac{1}{2}g_i(t)^2
r_{\vtheta,i}(\vx,t)
}_{\text{learned residual}}.
\label{eq:pf_split}
\end{equation}
The Gaussian component has the exact modewise propagator mapping a state at time \(\tau\) to time \(t\) via 
\begin{equation}
\phi_i(t,\tau)
 \: = \:
\sqrt{\frac{\gamma_i(t)}{\gamma_i(\tau)}}.
\label{eq:phi}
\end{equation}
Its derivation and composition property are given in Appendix~\ref{app:propagator}.

For a backward step \(t_1<t_0\) variation of constants gives
\begin{equation}
x_i(t_1)
\: = \:
\phi_i(t_1,t_0)x_i(t_0)
+
\int_{t_0}^{t_1}
\phi_i(t_1,\tau)
\left[-\frac{1}{2}g_i(\tau)^2\right]
r_{\vtheta,i}(\vx(\tau),\tau)\,d\tau.
\label{eq:solver}
\end{equation}
Thus, the Gaussian component is propagated exactly, while only the learned residual is approximated numerically.
The sampler uses two network evaluations per step and implementation details are given in Appendix~\ref{app:solver_details}.
This is related to the general strategy of diffusion-specific exponential integrators, which propagate analytically tractable components exactly while approximating the learned contribution numerically
\citep{lu2022dpmsolver,zhang2023deis}.
Here, the exact propagator is mode dependent and is induced by the fitted Gaussian reference.

\paragraph{Time grid.}
For the even NFE budgets used in our experiments each sampling step uses two network evaluations, so \(K=\mathrm{NFE}/2\). Following the power-law noise discretization of \citet{karras2022edm}, applied here to the conjugate noise scale \(q\) with \(q_{\min}=q(t_{\min})\) and
\(q_{\max}=q(t_{\max})\), we use
\begin{equation}
q_j
\: = \:
\left[
q_{\max}^{1/\rho}
+
\frac{j}{K}
\left(
q_{\min}^{1/\rho}
-
q_{\max}^{1/\rho}
\right)
\right]^\rho,
\qquad
j=0,\ldots,K.
\label{eq:grid}
\end{equation}
We then set \(t_j=q_j/\kappa\). Because \(q(t)=\kappa t\) is linear, \(\rho=1\) recovers the uniform grid used by the base exponential-residual sampler. 
We select a single \(\rho\) for all budgets by mean validation aMMD, with the uniform grid \(\rho=1\) as the baseline. This yields \(\rho=3\), which is fixed before any test evaluation. Unless a row is explicitly labelled \(\rho=1\), all reported exponential-residual results use \(\rho=3\) at every budget. Table~\ref{tab:grid} reports the corresponding test aMMD.

\paragraph{Heun sampler.}
``Heun'' denotes the deterministic second-order predictor--corrector solver \citep{ascher1998computer} for the probability-flow ODE. On the uniform-in-\(t\) grid used for all reported Heun results, each step evaluates the velocity at the current state, takes an Euler predictor step, and then evaluates the velocity again at the predicted state at the next time point. The two velocities are averaged for the corrected update. Thus, each step requires two network evaluations, so \(\mathrm{NFE}=2k\) corresponds to \(k\) sampling steps. The exponential-residual solver likewise uses two evaluations per step, so the two solvers are compared at equal NFE. EDM and the three preconditioning controls instead follow the EDM implementation, which omits the final correction (Appendix~\ref{app:samplers}).

\section{Experimental setup}
\label{sec:setup}

We evaluate five graph settings spanning three dataset families: METR-LA traffic speeds \citep{li2018traffic}, Molene temperature signals \citep{girault2015stationary}, and three synthetic stochastic block model (SBM) regimes with spectral concentration $c\in\{1,4,16\}$. The directed METR-LA adjacency matrix is symmetrized before constructing the graph Laplacian. We evaluate two terminal-noise regimes: $\kappa=2$ and $\kappa\approx32.6$ (targeting near-zero terminal SNR). All graph methods use the same backbone, optimization, batch size, random streams, and evaluation. In the central fitted-versus-scalar comparison the score model is fixed and only the Gaussian terminal initialization changes. Results are reported at the validation-converged stage, averaged over three random seeds.

Data generation quality is measured by aMMD \citep{gretton2012mmd}, the mean of the MMDs of quadratic variation, spectral centroid and degree correlation, based on the statistics and multi-bandwidth RBF estimator of \citet{rozada2026gad} on normalized signals. For a common evaluation operator, graph-based statistics are computed using the combinatorial Laplacian $\mD-\mW$, fixed across all methods.

We compare our proposed GRCD method against seven comparators: Elucidating Diffusion Models (EDM) \citep{karras2022edm}, two Whitened Score Diffusion (WSD)-style graph adaptations \citep{alido2025wsd}, and GAD \citep{rozada2026gad} under the matched graph protocol. GAD uses our matched protocol, so its reported aMMD values are not a reproduction of those in \citet{rozada2026gad}. The other three are preconditioning controls: isotropic, static graph, and dynamic conjugate preconditioning. Details are in Appendix~\ref{app:experimental_details}.

\section{Results}
\label{sec:results}
\subsection{Overall comparison}
\label{sec:overall}
Table~\ref{tab:main} shows the comparison on METR-LA.
At NFE~$4$ GRCD reaches an aMMD of $0.0538$.
Among external baselines EDM reaches an aMMD of $2.7757$, while the strongest method at this budget, the matched graph-prior WSD-style variant, reaches $1.2431$, which is still $23\times$ higher than GRCD.
The gap narrows with sampling budget: at NFE~$64$ the strongest external baseline or preconditioning control, i.e., conjugate preconditioning, reaches an aMMD of $0.0906$, while GRCD without the residual score reaches an aMMD of $0.0499$.
The practical advantage of our proposed method is therefore largest in the low-NFE regime.
Sampling cost is reported separately in Section~\ref{sec:efficiency}, where quality is plotted against measured wall-clock time.

\begin{table}[tbh]
\caption{\textbf{Matched comparison on METR-LA.}
aMMD at the validation-converged stage and a fixed nominal NFE budget (Appendix~\ref{app:samplers}).
Lower numbers are better and bold denotes
the best mean in each column. Results are mean $\pm$ sample standard
deviation over three random seeds. The last two rows use the exponential-residual solver.}
\label{tab:main}
\begin{center}
\small
\setlength{\tabcolsep}{5pt}
\begin{tabular}{@{}lcccc@{}}
\multicolumn{1}{c}{\bf METHOD} &
\multicolumn{1}{c}{\bf NFE 4} &
\multicolumn{1}{c}{\bf NFE 8} &
\multicolumn{1}{c}{\bf NFE 16} &
\multicolumn{1}{c}{\bf NFE 64}
\\ \hline \\
EDM
& $2.7757 \pm 0.0040$ & $1.3996 \pm 0.0605$ & $0.2640 \pm 0.0139$
& $0.1155 \pm 0.0053$
\\
GAD
& $2.9091 \pm 0.0016$ & $2.9093 \pm 0.0056$ & $2.8878 \pm 0.0146$
& $2.8785 \pm 0.0163$
\\
WSD-style, scalar
& $1.8752 \pm 0.2182$ & $0.3941 \pm 0.1355$ & $0.2695 \pm 0.0587$
& $0.2572 \pm 0.0183$
\\
WSD-style, graph prior
& $1.2431 \pm 0.2440$ & $0.2410 \pm 0.1347$ & $0.1980 \pm 0.0928$
& $0.0966 \pm 0.0325$
\\
Isotropic VE
& $2.7581 \pm 0.0084$ & $0.7810 \pm 0.0240$ & $0.1816 \pm 0.0139$
& $0.1152 \pm 0.0087$
\\
Static precond.
& $2.7728 \pm 0.0088$ & $0.7993 \pm 0.0155$ & $0.1755 \pm 0.0070$
& $0.1134 \pm 0.0043$
\\
Conjugate precond.
& $2.7581 \pm 0.0198$ & $0.6867 \pm 0.0344$ & $0.1385 \pm 0.0025$
& $0.0906 \pm 0.0016$
\\
GRCD w/o residual score
& $0.1463 \pm 0.0083$ & $0.0603 \pm 0.0029$ & $0.0513 \pm 0.0021$
& $\mathbf{0.0499 \pm 0.0018}$
\\
GRCD, $\rho{=}1$
& $0.5298 \pm 0.1303$ & $0.0519 \pm 0.0057$ & $\mathbf{0.0499 \pm 0.0032}$
& $0.0527 \pm 0.0032$
\\
GRCD, $\rho{=}3$
& $\mathbf{0.0538 \pm 0.0037}$ & $\mathbf{0.0503 \pm 0.0028}$ & $0.0509 \pm 0.0015$
& $0.0529 \pm 0.0028$
\\
\end{tabular}
\end{center}
\end{table}

GAD's results reflect a sampler run far outside its intended regime: at NFE~$4$, $96.5\%$ of its generated readings fall outside the $0$--$120$~mph range a loop detector can report. As a native-budget diagnostic we also ran it at the authors' own $1000$ and $3000$ steps, where it produces no out-of-range readings across three seeds; its aMMD there is $1.53 \pm 0.63$ and $1.81 \pm 0.63$, respectively, so the sampler is stable at those budgets but the variance across random seeds remains large.
Appendix~\ref{app:other-benchmark-results} repeats this comparison on the other four settings.
GRCD is best at NFE~$4$ on all of them by a factor of $22$--$36\times$, while the isotropic and static preconditioning controls and EDM overtake it at NFE~$64$ on the three SBM settings, as does the conjugate preconditioner on SBM~$c{=}16$.
Because the evaluation statistics are second order, we also compare GRCD with four non-learned Gaussian controls drawn directly from covariances fitted on the training split.
The best control consistently has a higher aMMD value than GRCD at NFE~$4$ on all five settings (cf.~Appendix~\ref{app:gaussian_control}).

\subsection{Terminal-reference effects across SNR regimes}
\label{sec:mechanism}

Section~\ref{sec:gaussian_reference} predicts that a fitted terminal reference matters only while the propagated terminal reference retains data-dependent spectral variance. As terminal SNR approaches zero, this dependence vanishes by \Eqref{eq:gmrf_limit_main}. 
NFE~$16$ is used for all mechanism comparisons in this paper: it is the smallest budget at which the Heun sampler shared by both arms has converged, so the contrast measures the terminal reference rather than discretization error.

\begin{figure}[tbh]
\begin{center}
\begin{minipage}[t]{0.49\linewidth}
\centering
\includegraphics[width=\linewidth]{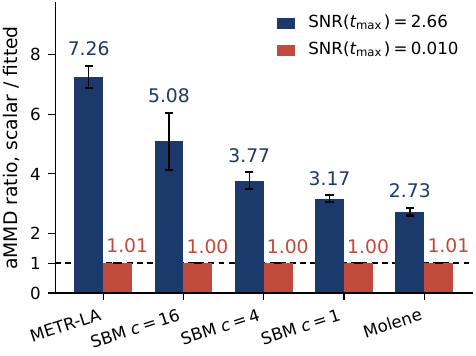}
\end{minipage}
\hfill
\begin{minipage}[t]{0.47\linewidth}
\centering
\includegraphics[width=\linewidth]{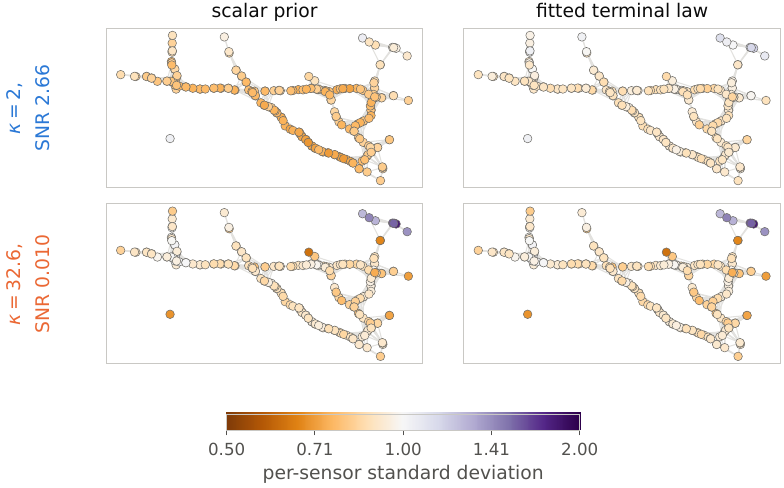}
\end{minipage}
\end{center}
\caption{\textbf{Terminal-reference effects are gated by terminal SNR.}
\textbf{Left:} Seed-paired scalar-to-fitted aMMD ratio at NFE~$16$ across five graph settings.
The dashed rule at $1$ marks no effect and values above it favor the fitted reference.
\textbf{Right:} Per-sensor standard deviation on METR-LA for one of the three random seeds with each dot placed at its sensor's longitude and latitude. Color is the generated standard deviation relative to the correct value of \(1.0\). Lower color contrast is better.}
\label{fig:terminal_snr}
\end{figure}

Figure~\ref{fig:terminal_snr} confirms this prediction empirically.
At $\kappa=2$ (i.e., terminal SNR $2.66$ on METR-LA) the fitted reference improves aMMD by $2.7\times$--$7.3\times$ across the five graph settings.
At $\kappa=32.6$ (i.e., terminal SNR $0.010$ on METR-LA) the ratios collapse to $1.00$--$1.01\times$.
Generated dispersion follows this pattern: on METR-LA the mean per-sensor standard deviation changes from $0.795$ to $0.913$ at finite SNR, but only from $0.916$ to $0.918$ at near-zero SNR.
Thus, the benefit of fitting the terminal reference vanishes as the propagated reference reaches near-zero SNR.

\subsection{Component contributions in GRCD}
\label{sec:ablation}

Table~\ref{tab:ladder} summarizes an ablation study that demonstrates the impact of GRCD's main components sequentially.
The dominant quality improvement comes from fitting the terminal reference: at NFE~$16$, aMMD drops from $0.3722$ to $0.0513$, which is a $7.25\times$ reduction.

\begin{table}[tb]
\caption{\textbf{Ablation ladder on METR-LA.}
aMMD at $\kappa=2$, validation-converged stage, mean $\pm$ s.d.~over three random seeds.
Each row adds one component to the row above. Lower numbers are better and bold denotes the best mean in each column. All rows use the uniform grid in $t$. The first and third rows are ablation-only configurations.}
\label{tab:ladder}
\begin{center}
\small
\setlength{\tabcolsep}{2.8pt}
\begin{tabular}{@{}lcccc@{}}
\multicolumn{1}{c}{\bf CONFIGURATION} &
\multicolumn{1}{c}{\bf NFE 8} &
\multicolumn{1}{c}{\bf NFE 12} &
\multicolumn{1}{c}{\bf NFE 16} &
\multicolumn{1}{c}{\bf NFE 64}
\\ \hline \\

Mode-dependent clock
& $0.4205 \pm 0.0085$
& $0.3817 \pm 0.0067$
& $0.3722 \pm 0.0069$
& $0.3632 \pm 0.0076$
\\

\quad + fitted terminal ref.
& $0.0603 \pm 0.0029$
& $0.0530 \pm 0.0023$
& $0.0513 \pm 0.0021$
& $\mathbf{0.0499 \pm 0.0018}$
\\

\quad + residual score
& $0.0622 \pm 0.0028$
& $0.0549 \pm 0.0034$
& $0.0537 \pm 0.0034$
& $0.0531 \pm 0.0033$
\\

\quad + exp. solver
& $\mathbf{0.0519 \pm 0.0057}$
& $\mathbf{0.0493 \pm 0.0036}$
& $\mathbf{0.0499 \pm 0.0032}$
& $0.0527 \pm 0.0032$
\\

\end{tabular}
\end{center}
\end{table}

The fitted terminal reference provides the main quality gain. The residual-score parameterization is structural rather than a direct improvement: at NFE~$16$ it changes aMMD from $0.0513$ to $0.0537$, but it separates the analytic Gaussian score from the learned residual and thereby enables the propagator in \Eqref{eq:phi}. The exponential-residual solver then shifts comparable quality toward smaller NFE.


\subsection{Solver and grid comparison}
\label{sec:solver_comparison}

The previous ablation study separates the fitted terminal reference from the numerical solver.
We therefore compare reverse integrators seed by seed with all solvers following identical evaluation pipelines.
NFE is counted from actual network evaluations, and every configuration below attains its requested budget exactly.
Heun and our exponential-residual solver use two evaluations per step, whereas DPM-Solver++ \citep{lu2025dpmsolverpp} multistep uses one.

For DPM-Solver++ we report all four tested configurations: $\{$2M, 3M$\}\times\{$uniform half-log-SNR grid, $\rho=3$ grid$\}$.
Our solver uses only the validation-selected $\rho=3$ grid.
DPM-Solver++ remains applicable because, by \Eqref{eq:snr_conditional}, the conditional SNR is identical across all graph-Fourier modes.
Its half-log-SNR coordinate
\(
\ell(t)
=
\frac{1}{2}\log \mathrm{SNR}^{\mathrm{cond}}_i(t)
=
-\log q(t)
\)
is therefore independent of $i$: the shrinkage factor $a_i(t)$ cancels between signal and conditional noise scales, so step size and extrapolation weights remain scalar.
This quantity is distinct from the mode-dependent reference terminal SNR in \Eqref{eq:snr_mode_main}.
The graph-frequency dependence enters the update only through $a_i(t)$.

The exponential-residual solver is competitive with DPM-Solver++ but does not outperform it consistently.
The strongest DPM-Solver++ configuration is 2M with the $\rho=3$ grid at every tested budget.
That configuration beats ours at NFE~$4$, while ours has lower means at NFE~$8$, $16$, and $32$.
The differences at larger budgets are small in absolute terms.
At NFE~$16$ and $32$ the mean differences from the best DPM-Solver++ configuration are $0.0010$ and $0.0003$.
The sampling grid has a larger effect at low NFE.
Replacing the uniform half-log-SNR grid with the validation-selected $\rho=3$ grid improves DPM-Solver++ at NFE~$4$ from $0.0659$ to $0.0523$ for 2M and from $0.0750$ to $0.0708$ for 3M.
For 2M, this improvement ($0.0136$) is over four times the difference between the best DPM-Solver++ configuration and our solver at the same budget. 
At NFE~$32$ the solver differences (\(\leq 0.0005\)) are an order of magnitude smaller than the shared-seed variation (\(\approx 0.003\)).

\begin{table}[tbh]
\caption{\textbf{Solver comparison.} aMMD on METR-LA, mean $\pm$ s.d.~over three random seeds. Lower numbers are better and \textbf{bold} marks the best mean at each NFE. DPM denotes DPM-Solver++.}
\label{tab:solvers}
\begin{center}
\small
\setlength{\tabcolsep}{2.8pt}
\begin{tabular}{@{}lcccc@{}}
\multicolumn{1}{c}{\bf SOLVER} &
\multicolumn{1}{c}{\bf NFE 4} &
\multicolumn{1}{c}{\bf NFE 8} &
\multicolumn{1}{c}{\bf NFE 16} &
\multicolumn{1}{c}{\bf NFE 32}
\\ \hline \\
Heun
& $0.1608 \pm 0.0205$
& $0.0606 \pm 0.0014$
& $0.0514 \pm 0.0027$
& $0.0523 \pm 0.0029$
\\
DPM 2M, uniform $\ell$
& $0.0659 \pm 0.0028$
& $0.0523 \pm 0.0022$
& $0.0509 \pm 0.0022$
& $0.0522 \pm 0.0029$
\\
DPM 2M, $\rho{=}3$
& $\mathbf{0.0523 \pm 0.0010}$
& $0.0516 \pm 0.0020$
& $0.0507 \pm 0.0026$
& $0.0522 \pm 0.0029$
\\
DPM 3M, uniform $\ell$
& $0.0750 \pm 0.0023$
& $0.0552 \pm 0.0021$
& $0.0516 \pm 0.0021$
& $0.0524 \pm 0.0029$
\\
DPM 3M, $\rho{=}3$
& $0.0708 \pm 0.0023$
& $0.0539 \pm 0.0022$
& $0.0510 \pm 0.0023$
& $0.0523 \pm 0.0029$
\\
Exp.\ residual ($\rho{=}3$, ours)
& $0.0552 \pm 0.0024$
& $\mathbf{0.0488 \pm 0.0028}$
& $\mathbf{0.0497 \pm 0.0020}$
& $\mathbf{0.0519 \pm 0.0029}$
\\
\end{tabular}
\end{center}
\end{table}
\subsection{Wall-clock efficiency}
\label{sec:efficiency}
\begin{figure}[tbh]
\begin{center}
\includegraphics[width=0.65\linewidth]{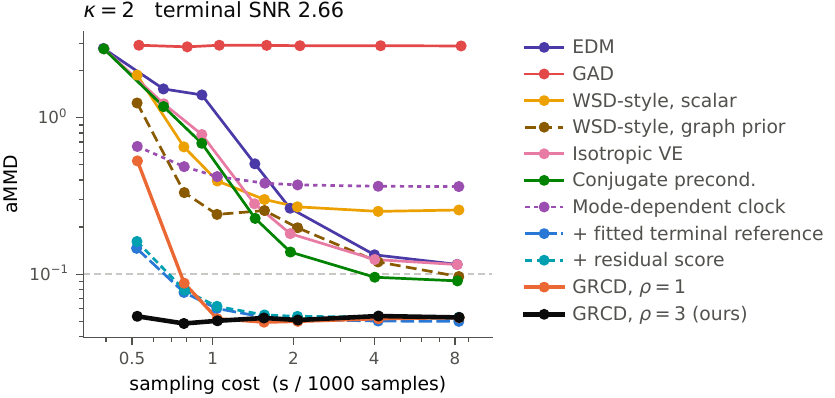}
\end{center}
\caption{\textbf{Wall-clock sampling efficiency on METR-LA.}
aMMD versus measured seconds per $1000$ generated samples. Lower and left are better.}
\label{fig:cost}
\end{figure}
NFE counts function evaluations, but methods with the same NFE can have different sampling times because non-network computations differ. Figure~\ref{fig:cost} therefore compares sample quality against measured wall-clock time on the same machine.
GRCD with the fitted terminal reference, residual score parameterization, exponential-residual solver, and validation-selected $\rho=3$ grid generates $1000$ samples in $0.5229$ seconds at NFE~$4$ and attains aMMD $0.0538$ there. No comparison method is close at that cost: the best at NFE~$4$ reaches $1.2431$, which is $23\times$ worse. The comparison methods become competitive only at NFE~$32$--$64$, where sampling costs $4.01$ (EDM at NFE~$32$) to $8.27$ (WSD at NFE~$64$) seconds per $1000$ samples, $7.7\times$--$15.8\times$ more than GRCD's
NFE-4 point. The main advantage is therefore that GRCD reaches useful sample quality much earlier on the wall-clock curve.
The exponential-residual sampler uses two function evaluations per reverse step, the same as Heun, so both pay the same price per step. Its wall-clock advantage comes from taking fewer steps: the analytic propagator absorbs the Gaussian part of each step, leaving the learned residual as the only piece integrated numerically.
GAD's curve is nearly flat because aMMD has saturated there. Its samples continue to change: between NFE~$4$ and NFE~$64$ the median per-sensor dispersion ratio falls from $166$ to $15$, an elevenfold drop, while aMMD moves from $2.9091$ to $2.8785$, so the metric registers only a small fraction of the change. We read the flat segment as a metric limit at large aMMD. 
Appendix~\ref{app:cost_target} reports the cheapest measured cost at which
each configuration reaches a fixed aMMD target. At the $0.1$ target GRCD
needs $0.52$ seconds against $4.02$ for the cheapest external baseline or preconditioning control that reaches it, a reduction of $1-0.52/4.02\approx 87\%$.

\section{Conclusion}
\label{sec:conclusion}
We study whether diffusion on structured signals needs to reach near-zero terminal SNR when a finite-SNR terminal reference can be estimated from the data.
GRCD uses a mode-dependent graph diffusion, fits a graph-spectral Gaussian reference from the training set and lets a neural network learn only the residual score.
At finite terminal SNR, the fitted reference improves generation because it preserves data-dependent spectral structure, a benefit that disappears as the terminal SNR approaches zero.
The same Gaussian reference also gives an exact modewise propagator, which enables accurate sampling with few function evaluations.

\paragraph{Limitations.}
Our construction assumes a fixed symmetric graph operator and loses directed structure, while the Gaussian reference models only second-order dependencies and permits invalid values for bounded signals. Performance gains peak in low-budget regimes, and alternative controls can outperform GRCD at higher NFE. For non-Gaussian data, a residual terminal-distribution mismatch may persist. Furthermore, the focus on unconditional generation leaves extensions to conditional tasks and higher-dimensional domains open.

\subsection*{Acknowledgements}
This work was supported by the Bayerisches Verbundforschungsprogramm (BayVFP) of the Free State of Bavaria under the funding line ``Digitalisierung''. We gratefully acknowledge this financial support.

\subsection*{AI use statement}
In this work, we used generative AI tools to improve the grammar and readability of the manuscript, identify potentially relevant literature, and assist with coding. All AI-assisted content was manually reviewed and verified. We take responsibility for the final content of this work.

\bibliography{iclr2027_conference}
\bibliographystyle{iclr2027_conference}

\appendix

\section{Appendix}
\subsection{Derivations for conjugate graph diffusion}
\label{app:conjugate_details}

\subsubsection{Derivation of mode equalization and the mode-dependent clock}
\label{app:conjugacy_proof}

Throughout this section, write
\[
\mL_\delta
=
\mU\operatorname{diag}(\evmu_1,\ldots,\evmu_N)\mU^\top.
\]
We use the same convention as in the main text: $\rvx_t$ denotes the graph-space process and $\widetilde{\rvx}_t$ its conjugate-coordinate representation. A subscript $i$ denotes the $i$-th graph-Fourier coefficient of the corresponding vector, e.g., $x_{t,i}:=[\mU^\top \rvx_t]_i$ and $\widetilde{x}_{t,i}:= [\mU^\top\widetilde{\rvx}_t]_i$. Likewise, $\epsilon_i:=[\mU^\top\rvepsilon]_i$.

We derive the shrinkage operator and the mode-dependent clock from the mode-equalization construction in Proposition~\ref{prop:conjugacy}. Consider one graph-Fourier mode with $\evmu_i>0$. We seek a linear graph-diffusion process of the form
\begin{equation}
d x_{t,i}
=
-\evmu_i c_i(t)x_{t,i}\,d t
+
\sqrt{2\sigma^2c_i(t)}\,d W_{t,i},
\label{eq:graph_sde_app}
\end{equation}
with $c_i(t)\geq0$, such that the change of coordinates
\begin{equation}
\widetilde{x}_{t,i}
=
\frac{x_{t,i}}{a_i(t)}
\label{eq:conjugate_mode_app}
\end{equation}
follows isotropic VE diffusion with noise amplitude $q(t)$:
\begin{equation}
d \widetilde{x}_{t,i}
=
\sqrt{2q(t)q'(t)}\,d W_{t,i}.
\label{eq:ve_sde_app}
\end{equation}
Here $a_i(0)=1$ and $q(0)=0$.

Writing
\[
x_{t,i}
=
a_i(t)\widetilde{x}_{t,i}
\]
and applying the product rule gives
\begin{equation}
d x_{t,i}
=
\frac{a_i'(t)}{a_i(t)}
x_{t,i}\,d t
+
a_i(t)\sqrt{2q(t)q'(t)}\,d W_{t,i}.
\label{eq:product_rule_app}
\end{equation}
Matching the drift and diffusion coefficients in \eqref{eq:graph_sde_app} and \eqref{eq:product_rule_app} yields
\begin{equation}
\frac{a_i'(t)}{a_i(t)}
=
-\evmu_i c_i(t),
\qquad
\sigma^2c_i(t)
=
q(t)q'(t)a_i(t)^2.
\label{eq:coefficient_matching_app}
\end{equation}
Substituting the second identity into the first gives
\begin{equation}
\frac{a_i'(t)}{a_i(t)}
=
-\frac{\evmu_i q(t)q'(t)}{\sigma^2}
a_i(t)^2.
\label{eq:a_mode_ode_app}
\end{equation}
Therefore
\begin{equation}
\frac{d}{d t}a_i(t)^{-2}
=
\frac{2\evmu_i q(t)q'(t)}{\sigma^2}
=
\frac{\evmu_i}{\sigma^2}
\frac{d}{d t}q(t)^2.
\label{eq:a_inverse_ode_app}
\end{equation}
Integrating from $0$ to $t$ and using $a_i(0)=1$ and $q(0)=0$ gives
\begin{equation}
a_i(t)^{-2}
=
1+\frac{\evmu_iq(t)^2}{\sigma^2},
\end{equation}
hence
\begin{equation}
a_i(t)
=
\left(
1+\frac{\evmu_iq(t)^2}{\sigma^2}
\right)^{-1/2}.
\label{eq:ai_derivation_app}
\end{equation}

Since
\[
\mL_\delta
=
\mU\operatorname{diag}(\evmu_1,\ldots,\evmu_N)\mU^\top,
\]
spectral functional calculus gives
\begin{equation}
\mA_t
=
\mU
\operatorname{diag}\!\left(
a_1(t),\ldots,a_N(t)
\right)
\mU^\top
=
\left(
\mI+\frac{q(t)^2}{\sigma^2}\mL_\delta
\right)^{-1/2},
\label{eq:At_derivation_app}
\end{equation}
which is exactly the shrinkage operator in \eqref{eq:At}.

Finally, substituting \eqref{eq:ai_derivation_app} into the diffusion coefficient identity in \eqref{eq:coefficient_matching_app} gives
\begin{equation}
c_i(t)
=
\frac{q(t)q'(t)}
{\sigma^2+\evmu_iq(t)^2},
\label{eq:ci_derivation_app}
\end{equation}
which is the mode-dependent clock in \eqref{eq:Mode-dependent_clock}.

\paragraph{Modewise conditional marginals.}
Equation~\eqref{eq:forward} acts in graph-Fourier mode $i$ as
\begin{equation}
x_{t,i}
=
a_i(t)
\left(
x_{0,i}
+
q(t)\epsilon_i
\right),
\qquad
\epsilon_i
\sim
\mathcal{N}(0,1).
\label{eq:forward_mode_app}
\end{equation}
Since
\[
\widetilde{x}_{t,i}
=
a_i(t)^{-1}x_{t,i},
\]
the conjugate coordinates satisfy
\[
\widetilde{x}_{t,i}
=
x_{0,i}
+
q(t)\epsilon_i.
\]
Because $\mA_0=\mI$, we also have $\widetilde{\rvx}_0=\rvx_0$. Therefore
\begin{equation}
\Var\!\left(
\widetilde{x}_{t,i}
\mid
\widetilde{x}_{0,i}
\right)
=
q(t)^2
\qquad
\text{for every mode }i.
\label{eq:equalise_app}
\end{equation}

\paragraph{Uniqueness within the linear graph-diffusion family.}
For a prescribed differentiable nondecreasing \(q(t)\) with \(q(0)=0\), coefficient matching in \eqref{eq:coefficient_matching_app}, together with $a_i(0)=1$, uniquely determines both $a_i(t)$ and $c_i(t)$. Equivalently, the integrated mode-dependent clock is
\begin{equation}
C_i(t)
:=
\int_0^t c_i(u)\,d u
=
\frac{1}{2\evmu_i}
\log\!\left(
1+\frac{\evmu_iq(t)^2}{\sigma^2}
\right).
\label{eq:integrated_clock_app}
\end{equation}
Thus the mode dependence is induced by exact conjugacy to the prescribed isotropic VE corruption within this linear graph-diffusion family.

\subsubsection{Why a shared scalar clock cannot exactly equalize modes}
\label{app:shared_clock}

To isolate the role of the mode-dependent clock, consider instead a single shared scalar clock $c(t)$ for every graph-Fourier mode:
\begin{equation}
d x^{\mathrm{sh}}_{t,i}
=
-\evmu_i c(t)x^{\mathrm{sh}}_{t,i}\,d t
+
\sqrt{2\sigma^2c(t)}\,d W_{t,i}.
\label{eq:shared_sde_app}
\end{equation}
Let
\begin{equation}
C(t)
=
\int_0^t c(u)\,d u.
\label{eq:shared_integrated_clock_app}
\end{equation}
The deterministic attenuation of mode $i$ is
\[
a_i^{\mathrm{sh}}(t)
=
\exp\!\left(
-\evmu_iC(t)
\right),
\]
and its conditional variance in the native graph-Fourier coordinates is
\begin{equation}
\Var\!\left(
x^{\mathrm{sh}}_{t,i}
\mid
x^{\mathrm{sh}}_{0,i}
\right)
=
\frac{\sigma^2}{\evmu_i}
\left[
1-
\exp\!\left(
-2\evmu_iC(t)
\right)
\right].
\label{eq:shared_native_variance_app}
\end{equation}
Undoing the deterministic attenuation defines the transformed shared-clock coordinate
\begin{equation}
\widetilde{x}^{\mathrm{sh}}_{t,i}
=
\frac{x^{\mathrm{sh}}_{t,i}}{a_i^{\mathrm{sh}}(t)}.
\label{eq:shared_conjugate_coordinate_app}
\end{equation}
Its conditional variance in these coordinates is
\begin{equation}
\Var\!\left(
\widetilde{x}^{\mathrm{sh}}_{t,i}
\mid
\widetilde{x}^{\mathrm{sh}}_{0,i}
\right)
=
\frac{\sigma^2}{\evmu_i}
\left[
\exp\!\left(
2\evmu_iC(t)
\right)
-
1
\right].
\label{eq:shared_conjugate_variance_app}
\end{equation}
Unlike \eqref{eq:equalise_app}, this expression depends explicitly on
$\evmu_i$.

Matching the common target variance $q(t)^2$ in mode $i$ would require
\begin{equation}
C_i^\star(t)
=
\frac{1}{2\evmu_i}
\log\!\left(
1+\frac{\evmu_iq(t)^2}{\sigma^2}
\right).
\label{eq:shared_matching_clock_app}
\end{equation}
For fixed $t$ with $q(t)>0$, this quantity is strictly decreasing in
$\evmu_i$. To see this, let
\[
b
=
\frac{q(t)^2}{\sigma^2}
>
0.
\]
Then
\begin{equation}
\frac{d}{d \evmu_i}
C_i^\star(t)
=
\frac{
\frac{b\evmu_i}{1+b\evmu_i}
-
\log(1+b\evmu_i)
}{
2\evmu_i^2
}.
\end{equation}
For $x>0$,
\begin{equation}
\log(1+x)
-
\frac{x}{1+x}
=
\int_0^x
\frac{u}{(1+u)^2}\,d u
>
0,
\end{equation}
so the derivative is strictly negative. Hence, for two graph modes with distinct eigenvalues, a single shared scalar clock cannot realize the common conjugate variance $q(t)^2$ in both modes, which is why the construction requires a mode-dependent clock.

\subsection{Gaussian-reference details}
\label{app:reference}
\subsubsection{Spectral diagonal approximation}

The reference used in the main text is obtained from the Ledoit--Wolf covariance
\(\mSigma_{\mathrm{LW}}\) fitted on the training split. Before retaining only
its diagonal, define the full covariance in the graph-Fourier basis as
\begin{equation}
\widetilde{\mSigma}
=
\mU^\top
\left(
\mSigma_{\mathrm{LW}}+\varepsilon_{\mathrm{ref}}\mI
\right)
\mU.
\label{eq:project_cov_app}
\end{equation}
The reference retains only its diagonal entries,
\begin{equation}
\evv_i
=
\widetilde{\mSigma}_{ii},
\qquad
\mSigma_{\mathrm{ref}}
=
\mU
\operatorname{diag}(\vv)
\mU^\top.
\label{eq:reference_cov_app}
\end{equation}
Thus, cross-mode covariances are discarded. The discarded off-diagonal energy,
\(\|\widetilde{\mSigma}-\operatorname{diag}(\widetilde{\mSigma})\|_F/
\|\widetilde{\mSigma}\|_F\), 
is \(0.6677\) on METR-LA, so the reference is a deliberately coarse covariance model whose remaining discrepancy is left to the learned residual.

\subsubsection{Exact propagated marginal of the Gaussian reference}

Suppose, for this derivation only, that
\begin{equation}
\rvx_0
\sim
\mathcal{N}\!\left(
\vzero,\mSigma_{\mathrm{ref}}
\right).
\label{eq:reference_assumption_app}
\end{equation}
In graph-Fourier mode $i$, \eqref{eq:forward} gives
\begin{equation}
x_{t,i}
=
a_i(t)
\left(
x_{0,i}
+
q(t)\epsilon_i
\right).
\end{equation}
Since
\[
x_{0,i}
\sim
\mathcal{N}(0,\evv_i),
\qquad
\epsilon_i
\sim
\mathcal{N}(0,1),
\]
and the two are independent,
\begin{align}
\gamma_i(t)
&=
\Var\!\left(
x_{t,i}
\right)
\\
&=
a_i(t)^2
\left(
\evv_i+q(t)^2
\right)
\\
&=
\frac{
\sigma^2
\left(
\evv_i+q(t)^2
\right)
}{
\sigma^2+\evmu_iq(t)^2
}.
\label{eq:gamma_derivation_app}
\end{align}
Therefore
\begin{equation}
p_{\mathrm{ref},t}
=
\mathcal{N}\!\left(
\vzero,\,
\mU
\operatorname{diag}\!\left(
\gamma_1(t),\ldots,\gamma_N(t)
\right)
\mU^\top
\right),
\end{equation}
which proves \eqref{eq:gamma} for the propagated Gaussian reference.

\subsubsection{Reference marginal versus the true data marginal}

The Gaussian calculation above does not imply that the true marginal $p_t$ is Gaussian. For an arbitrary initial data distribution $p_0$, the conjugate marginal is
\begin{equation}
\widetilde p_t
=
p_0
*
\mathcal{N}\!\left(
\vzero,q(t)^2\mI
\right),
\label{eq:true_conjugate_marginal_app}
\end{equation}
and the graph-space marginal is its pushforward under $\mA_t$:
\begin{equation}
p_t
=
(\mA_t)_{\#}
\left[
p_0
*
\mathcal{N}\!\left(
\vzero,q(t)^2\mI
\right)
\right].
\label{eq:true_marginal_app}
\end{equation}
This distribution is generally non-Gaussian. Hence $p_{\mathrm{ref},t}$ is the exact propagated marginal of the chosen Gaussian reference, but is not identified with the true data marginal unless the initial data distribution actually equals the Gaussian reference.

\subsection{Residual-score target derivation}
\label{app:residual_target}

Let
\begin{equation}
\eta_i(t)
=
q(t)a_i(t)
\label{eq:eta_app}
\end{equation}
denote the conditional standard deviation in graph-Fourier mode \(i\). From
\eqref{eq:forward},
\begin{equation}
x_{t,i}
-
a_i(t)x_{0,i}
=
\eta_i(t)\epsilon_i.
\label{eq:conditional_noise_app}
\end{equation}
Hence the conditional distribution is Gaussian with score
\begin{align}
\frac{\partial}{\partial x_{t,i}}
\log
p\!\left(
x_{t,i}
\mid
x_{0,i}
\right)
&=
-
\frac{
x_{t,i}
-
a_i(t)x_{0,i}
}{
\eta_i(t)^2
}
\\
&=
-
\frac{
\epsilon_i
}{
\eta_i(t)
}.
\label{eq:conditional_score_app}
\end{align}

The propagated Gaussian reference has graph-Fourier score
\begin{equation}
s_{\mathrm{ref},i}(\vx,t)
=
-\frac{x_i}{\gamma_i(t)},
\qquad
x_i=[\mU^\top\vx]_i.
\label{eq:reference_score_app}
\end{equation}
Let \(s_i^*(\vx,t)\) denote the true marginal score in mode \(i\), and define
\begin{equation}
r_i^*(\vx,t)
=
s_i^*(\vx,t)
-
s_{\mathrm{ref},i}(\vx,t).
\label{eq:true_residual_split_app}
\end{equation}
Equivalently,
\begin{equation}
r_i^*(\vx,t)
=
\left[
\mU^\top\nabla_{\vx}
\log
\frac{
p_t(\vx)
}{
p_{\mathrm{ref},t}(\vx)
}
\right]_i,
\label{eq:density_ratio_score_app}
\end{equation}
so the residual is the \(i\)-th graph-Fourier coefficient of the density-ratio score between the true marginal and the propagated Gaussian reference.

Let \(f_{\vtheta}(\vx,t)\) denote the network output in the node basis and
\[
f_{\vtheta,i}(\vx,t)
=
[\mU^\top f_{\vtheta}(\vx,t)]_i
\]
its \(i\)-th graph-Fourier coefficient. We parameterize the residual as
\begin{equation}
r_{\vtheta,i}(\vx,t)
=
-\frac{
f_{\vtheta,i}(\vx,t)
}{
\eta_i(t)
}.
\label{eq:residual_param_app}
\end{equation}
The reconstructed score is therefore
\begin{equation}
s_{\vtheta,i}(\vx,t)
=
s_{\mathrm{ref},i}(\vx,t)
-
\frac{
f_{\vtheta,i}(\vx,t)
}{
\eta_i(t)
}.
\label{eq:reconstructed_score_app}
\end{equation}

Using the conditional score in
\eqref{eq:conditional_score_app} as the denoising target gives
\begin{equation}
s_{\mathrm{ref},i}(\rvx_t,t)
-
\frac{
f_i^*
}{
\eta_i(t)
}
=
-\frac{
\epsilon_i
}{
\eta_i(t)
}.
\end{equation}
Multiplying by \(\eta_i(t)\) yields
\begin{align}
f_i^*
&=
\epsilon_i
+
\eta_i(t)
s_{\mathrm{ref},i}(\rvx_t,t)
\\
&=
\epsilon_i
-
\eta_i(t)
\frac{
x_{t,i}
}{
\gamma_i(t)
}.
\label{eq:target_derivation_app}
\end{align}
This is the target used in \eqref{eq:loss}.

\paragraph{Consistency with the marginal score.}
Under the squared loss in \eqref{eq:loss}, the population minimizer is the conditional expectation of the target:
\begin{align}
f_i^{\mathrm{opt}}(\vx,t)
&=
\E\!\left[
f_i^*
\mid
\rvx_t=\vx
\right]
\\
&=
\E\!\left[
\epsilon_i
\mid
\rvx_t=\vx
\right]
-
\eta_i(t)
\frac{x_i}{\gamma_i(t)}.
\label{eq:optimal_network_app}
\end{align}
The marginal score satisfies
\begin{equation}
s_i^*(\vx,t)
=
\E\!\left[
\left[
\mU^\top
\nabla_{\rvx_t}
\log p(\rvx_t\mid\rvx_0)
\right]_i
\;\middle|\;
\rvx_t=\vx
\right]
=
-\frac{
\E[\epsilon_i\mid\rvx_t=\vx]
}{
\eta_i(t)
}.
\label{eq:marginal_score_identity_app}
\end{equation}

Substituting \eqref{eq:optimal_network_app} into \eqref{eq:reconstructed_score_app} gives
\begin{align}
s_{\vtheta,i}(\vx,t)
=
s_{\mathrm{ref},i}(\vx,t)
-
\frac{
f_i^{\mathrm{opt}}(\vx,t)
}{
\eta_i(t)
}
&=
-\frac{x_i}{\gamma_i(t)}
-
\frac{
\E[\epsilon_i\mid\rvx_t=\vx]
}{
\eta_i(t)
}
+
\frac{x_i}{\gamma_i(t)}
\\
&=
-\frac{
\E[\epsilon_i\mid\rvx_t=\vx]
}{
\eta_i(t)
}
\\
&=
s_i^*(\vx,t).
\end{align}
Thus, at the population optimum, the residual parameterization exactly recovers the marginal score.

If the Gaussian reference score is disabled,
\(s_{\mathrm{ref},i}\equiv0\), then
\begin{equation}
f_i^*
=
\epsilon_i,
\end{equation}
recovering the corresponding standard \(\epsilon\)-prediction objective.

\paragraph{Gaussian-reference exactness.}
If the data distribution equals the fitted Gaussian reference at \(t=0\), then
\begin{equation}
p_t
=
p_{\mathrm{ref},t}
\end{equation}
for every \(t\), and therefore
\begin{equation}
r_i^*(\vx,t)
=
0
\end{equation}
for every mode. In this special case, the Gaussian reference supplies the complete marginal score analytically and no learned residual is required.

\subsection{Derivation of the exact Gaussian propagator}
\label{app:propagator}

For the modewise forward SDE, define
\begin{equation}
b_i(t)
=
-\evmu_i c_i(t),
\qquad
g_i(t)^2
=
2\sigma^2c_i(t).
\label{eq:bg_app}
\end{equation}
The probability-flow ODE is
\begin{equation}
\frac{d x_i}{d t}
=
b_i(t)x_i
-
\frac{1}{2}
g_i(t)^2
s_{\vtheta,i}(\vx,t).
\label{eq:pf_ode_app}
\end{equation}
Substituting
\[
s_{\vtheta,i}(\vx,t)
=
-\frac{x_i}{\gamma_i(t)}
+
r_{\vtheta,i}(\vx,t)
\]
gives
\begin{equation}
\frac{d x_i}{d t}
=
\left[
b_i(t)
+
\frac{
g_i(t)^2
}{
2\gamma_i(t)
}
\right]
x_i
-
\frac{1}{2}
g_i(t)^2
r_{\vtheta,i}(\vx,t).
\label{eq:pf_split_app}
\end{equation}

Because $\gamma_i(t)$ is the variance of the Gaussian reference propagated under the linear SDE, it satisfies the standard linear variance equation
\begin{equation}
\frac{d\gamma_i(t)}{d t}
=
2b_i(t)\gamma_i(t)
+
g_i(t)^2.
\label{eq:gamma_ode_app}
\end{equation}
Dividing by $2\gamma_i(t)$ gives
\begin{equation}
b_i(t)
+
\frac{
g_i(t)^2
}{
2\gamma_i(t)
}
=
\frac{1}{2}
\frac{d}{d t}
\log\gamma_i(t).
\label{eq:linear_log_gamma_app}
\end{equation}

The exact propagator of the Gaussian component from time $s$ to time $t$ is therefore
\begin{align}
\phi_i(t,s)
&=
\exp\!\left(
\int_s^t
\left[
b_i(u)
+
\frac{
g_i(u)^2
}{
2\gamma_i(u)
}
\right]
\,d u
\right)
\\
&=
\exp\!\left(
\frac{1}{2}
\int_s^t
\frac{d}{d u}
\log\gamma_i(u)
\,d u
\right)
\\
&=
\sqrt{
\frac{
\gamma_i(t)
}{
\gamma_i(s)
}
}.
\label{eq:phi_derivation_app}
\end{align}
This is exactly \eqref{eq:phi}.

Substituting the closed-form reference variance from \eqref{eq:gamma} gives
\begin{equation}
\phi_i(t,s)
=
\sqrt{
\frac{
\evv_i+q(t)^2
}{
\evv_i+q(s)^2
}
}
\sqrt{
\frac{
\sigma^2+\evmu_iq(s)^2
}{
\sigma^2+\evmu_iq(t)^2
}
}.
\label{eq:phi_explicit_app}
\end{equation}

The propagator has the exact composition property
\begin{align}
\phi_i(t,u)\phi_i(u,s)
&=
\sqrt{
\frac{\gamma_i(t)}{\gamma_i(u)}
}
\sqrt{
\frac{\gamma_i(u)}{\gamma_i(s)}
}
\\
&=
\sqrt{
\frac{\gamma_i(t)}{\gamma_i(s)}
}
\\
&=
\phi_i(t,s).
\label{eq:phi_composition_app}
\end{align}
Thus the linear Gaussian component is propagated exactly between arbitrary
time points.

If
$r_{\vtheta,i}\equiv0$, then \eqref{eq:pf_split_app} reduces to the known
linear Gaussian flow and
\begin{equation}
x_i(t)
=
\phi_i(t,s)x_i(s)
\end{equation}
exactly. The exponential-residual sampler is therefore exact for the probability-flow dynamics in the zero-residual Gaussian-reference case.

\subsection{Exponential-residual solver details}
\label{app:solver_details}
For a backward interval \(t_1<t_0\), write the residual integral in
\Eqref{eq:solver} using
\begin{equation}
B_i(\tau;t_1)
=
-\frac{1}{2}
\phi_i(t_1,\tau)g_i(\tau)^2,
\qquad
\alpha(\tau)
=
\frac{\tau-t_0}{t_1-t_0}.
\label{eq:B_alpha_app}
\end{equation}
We approximate the learned residual linearly between its endpoint values,
\begin{equation}
r_{\vtheta,i}(\vx(\tau),\tau)
\approx
(1-\alpha(\tau))r_{0,i}
+
\alpha(\tau)r_{1,i},
\end{equation}
with corresponding weights
\begin{equation}
w_{0,i}
=
\int_{t_0}^{t_1}
B_i(\tau;t_1)(1-\alpha(\tau))\,d\tau,
\qquad
w_{1,i}
=
\int_{t_0}^{t_1}
B_i(\tau;t_1)\alpha(\tau)\,d\tau.
\label{eq:residual_weights_app}
\end{equation}

At the beginning of the step,
\[
r_{0,i}
=
r_{\vtheta,i}(\vx(t_0),t_0).
\]
Using \(r_{0,i}\) at both endpoints gives the predictor
\begin{equation}
x_{i,\mathrm{pred}}
=
\phi_i(t_1,t_0)x_i(t_0)
+
(w_{0,i}+w_{1,i})r_{0,i}.
\label{eq:predictor_app}
\end{equation}
We then evaluate
\[
r_{1,i}
=
r_{\vtheta,i}(\vx_{\mathrm{pred}},t_1),
\]
where \(\vx_{\mathrm{pred}}\) denotes the signal with graph-Fourier coefficients \(x_{i,\mathrm{pred}}\), and apply the corrected update
\begin{equation}
x_i(t_1)
=
\phi_i(t_1,t_0)x_i(t_0)
+
w_{0,i}r_{0,i}
+
w_{1,i}r_{1,i}.
\label{eq:corrector_app}
\end{equation}

\subsection{Experimental details}
\label{app:experimental_details}

\subsubsection{Datasets and splits}
\label{app:datasets}

Table~\ref{tab:app_datasets} summarizes the five graph-signal settings. The two real datasets use chronological splits, while the synthetic SBM datasets use independently generated train, validation and test sets. All normalization, covariance fitting and reference estimation use the training split only.

The chronological rule is $70/10/20$ on both real datasets. On Molene it gives $520/74/150$ signals after integer rounding. On METR-LA it gives $23{,}990/3{,}427/6{,}855$, which the caps listed in Table~\ref{tab:app_hparams} then reduce to the sizes shown, by evenly spaced subsampling so that each split still spans its full time period. This is why the METR-LA row is not exactly $70/10/20$.

\begin{table}[t]
\caption{\textbf{Datasets.}
$N$ denotes the number of graph nodes. Split sizes are the numbers of graph signals used in each experiment.}
\label{tab:app_datasets}
\begin{center}
\small
\begin{tabular}{@{}llrrrr@{}}
\multicolumn{1}{c}{\bf SETTING} &
\multicolumn{1}{c}{\bf SOURCE} &
\multicolumn{1}{c}{\bf $N$} &
\multicolumn{1}{c}{\bf TRAIN} &
\multicolumn{1}{c}{\bf VAL} &
\multicolumn{1}{c}{\bf TEST}
\\ \hline \\
METR-LA        & traffic, $5$-min \citep{li2018traffic}
               & $207$ & $20{,}000$ & $3{,}000$ & $5{,}000$ \\
Molene         & temperature \citep{girault2015stationary}
               & $37$  & $520$      & $74$      & $150$ \\
SBM ($c{=}1$)  & synthetic
               & $32$  & $4{,}000$  & $1{,}000$ & $5{,}000$ \\
SBM ($c{=}4$)  & synthetic
               & $32$  & $4{,}000$  & $1{,}000$ & $5{,}000$ \\
SBM ($c{=}16$) & synthetic
               & $32$  & $4{,}000$  & $1{,}000$ & $5{,}000$ \\
\end{tabular}
\end{center}
\end{table}

\paragraph{METR-LA.}
We use the adjacency matrix distributed with DCRNN \citep{li2018traffic}. The source graph is strongly directional: $1{,}111$ of its $1{,}313$ edges occur in one direction only and carry $75.0\%$ of the total edge weight. 
We therefore symmetrize the graph by the half-sum for the common protocol.
Repeating the exponential-residual arm and the conjugate preconditioning control under $\max(\mW,\mW^\top)$ leaves the ordering unchanged: at NFE~$16$ the former moves from $0.0499$ to $0.0536$ and the latter from $0.1385$ to $0.1295$, so the gap between them is preserved under either symmetrizer.

\paragraph{Molene.}
The dataset contains $744$ hourly temperature measurements from $37$ stations in Brittany. We build a $5$-nearest-neighbor Gaussian graph from the station coordinates, with bandwidth equal to the median pairwise distance. The original Molene study does not define a train/validation/test split for generative modeling, so we apply the same chronological rule as for METR-LA. The resulting $150$ test signals make this the smallest evaluation set in the study, so its aMMD estimates carry visibly larger seed-to-seed spread than the $5{,}000$-sample settings (Table~\ref{tab:app_molene}).

\paragraph{SBM.}
We generate a connected two-block stochastic block model with $N=32$,
$p_{\rm in}=0.4$ and $p_{\rm out}=0.04$.
Signals are sampled in the eigenbasis of its combinatorial Laplacian
\(\mL_{\rm comb}\), with spectral variance
\[
v_0(\nu_i)=0.2+\frac{0.8}{1+c\,\nu_i},
\qquad
c\in\{1,4,16\},
\]
together with a bimodal $\pm3$ shift in the second graph-Fourier mode, the
Fiedler vector, which separates the two blocks. Here
\(\nu_i=\lambda_i(\mL_{\rm comb})/\lambda_{\max}(\mL_{\rm comb})+\delta
\in[\delta,1+\delta]\) are the scaled and shifted eigenvalues of
\(\mL_{\rm comb}\).
The parameter $c$ controls how sharply signal energy concentrates at low graph
frequencies, while the bimodal component makes the data non-Gaussian, so that
its distribution cannot be represented exactly by a covariance-matched
Gaussian. The graph and the dataset are generated once and held fixed. The
three experimental seeds vary only model initialization, batch order and
sampling noise.

\subsubsection{Training protocol}
\label{app:backbone}
All methods in the common protocol use the \texttt{DenoiserGNN} architecture released with GAD \citep{rozada2026gad}, with three $K=5$ graph-filter blocks and a $64$-dimensional time embedding. In our common protocol, we set the hidden width to $128$. Training uses Adam with the hyperparameters in Table~\ref{tab:app_hparams}. Validation denoising error is evaluated every $250$ updates using EMA weights, and training stops after eight consecutive evaluations without a relative improvement greater than $0.2\%$, subject to a maximum of $40{,}000$ updates. All methods therefore share one stopping criterion. No arm reaches the $40{,}000$-update cap.

For paired comparisons, configurations with the same experimental seed receive the same batches, forward-noise draws, and time draws at each update.

\subsubsection{Noise regimes}
\label{app:forward}

GRCD and the WSD-style adaptations use
\[
q(t)=\kappa t,
\qquad
t\in[0.02,1].
\]
The finite-SNR setting uses $\kappa=2$, giving $q\in[0.04,2.0]$ and an energy-weighted terminal SNR of $2.663$ on METR-LA. The near-zero-SNR setting uses $\kappa=32.6356$ on METR-LA, calibrated to terminal SNR $0.010$, and $\kappa=32.64$ on the other settings, where the resulting terminal SNR ranges from $0.003$ to $0.027$. On all five settings this is two to three orders of magnitude below the corresponding finite-SNR value.

The isotropic VE controls use
\[
y_t=x_0+q\rvepsilon,
\qquad
q\in[0.02,32.64],
\]
with $P_{\rm mean}=-1.2$, $P_{\rm std}=1.2$ during training and a $\rho_{\rm VE}=7$ power-law grid during sampling.

\subsubsection{Baseline implementation}
\label{app:baseline_config}

EDM \citep{karras2022edm} uses its published preconditioning, loss weighting, noise schedule, and second-order Heun sampler, with the shared graph backbone substituted for the image network. We use $P_{\rm mean}=-1.2$, $P_{\rm std}=1.2$, $\sigma_{\min}=0.002$, $\sigma_{\max}=80$, and $\rho=7$. GAD \citep{rozada2026gad} uses its published graph-aware process and native Euler--Maruyama sampler. Following the authors' released implementation, we use two Newton iterations to invert the heat-time map and a sampling grid descending to $s=0$.

Trained instead at the authors' own configuration ($3{,}985$ parameters, $5{,}000$ epochs, no early stopping) and sampled at their evaluation script's $5{,}000$ steps, GAD \citep{rozada2026gad} reaches aMMD $2.10 \pm 0.15$ over three seeds, so it is not better under its own settings than under ours. That configuration costs $785{,}000$ optimizer updates, against at most $25{,}000$ for any arm in the matched protocol, and sampling at its native budget takes $261$ seconds per $1000$ samples against GRCD's $0.52$.

For the WSD-style adaptations \citep{alido2025wsd}, we transfer the whitened-score parameterization to the conjugate forward process and evaluate both a scalar variance and the same Ledoit--Wolf graph-spectral covariance used by GRCD.

The isotropic, static-graph, and dynamic-conjugate preconditioning controls use the same isotropic VE corruption and deterministic Heun sampler, differing only in the preconditioner.

\subsubsection{Sampler details}
\label{app:samplers}

The exponential-residual solver propagates the Gaussian component analytically and evaluates the residual integral with composite Simpson quadrature on $65$ nodes after linear interpolation of the residual in time. The quadrature adds no network evaluations, so each reverse step uses two evaluations.
Following the reference EDM Heun implementation, a $k$-step trajectory uses
$2k-1$ network evaluations because the second-order correction is omitted on
the final step to zero noise. On our nominal even-NFE grid, EDM and the three
Heun preconditioning controls therefore use one evaluation fewer than the
nominal budget. All other configurations use the nominal budget exactly.

Reverse steps use
\begin{equation}
q_j
=
\left(
q_{\max}^{1/\rho}
+
\frac{j}{K}
\left(q_{\min}^{1/\rho}-q_{\max}^{1/\rho}\right)
\right)^\rho,
\qquad
j=0,\ldots,K.
\label{eq:app_rho_grid}
\end{equation}
The base sampler uses $\rho=1$. We select $\rho=3$ once on the METR-LA validation split and use it unchanged for all sampling budgets and datasets. Table~\ref{tab:grid} reports the corresponding test aMMD.

\begin{table}[ht]
\caption{\textbf{Time-grid sweep for the exponential-residual solver on
METR-LA.} Test aMMD at $\kappa=2$, converged, mean $\pm$ standard deviation over three random seeds.}
\label{tab:grid}
\begin{center}
\small
\setlength{\tabcolsep}{4pt}
\begin{tabular}{@{}lccccc@{}}
\multicolumn{1}{c}{\bf GRID} &
\multicolumn{1}{c}{\bf NFE 4} &
\multicolumn{1}{c}{\bf NFE 6} &
\multicolumn{1}{c}{\bf NFE 8} &
\multicolumn{1}{c}{\bf NFE 12} &
\multicolumn{1}{c}{\bf NFE 16}
\\ \hline \\
$\rho=1$ 
& $0.5298 \pm 0.1303$ & $0.0876 \pm 0.0257$ & $0.0519 \pm 0.0057$
& $0.0493 \pm 0.0036$ & $0.0499 \pm 0.0032$
\\
$\rho=2$
& $0.0702 \pm 0.0127$ & $0.0482 \pm 0.0019$ & $0.0498 \pm 0.0030$
& $0.0522 \pm 0.0015$ & $0.0508 \pm 0.0016$
\\
$\rho=3$ 
& $0.0538 \pm 0.0037$ & $0.0484 \pm 0.0011$ & $0.0503 \pm 0.0028$
& $0.0525 \pm 0.0015$ & $0.0509 \pm 0.0015$
\\
$\rho=5$
& $0.0558 \pm 0.0028$ & $0.0482 \pm 0.0006$ & $0.0504 \pm 0.0027$
& $0.0525 \pm 0.0016$ & $0.0509 \pm 0.0015$
\\
$\rho=7$
& $0.0589 \pm 0.0031$ & $0.0482 \pm 0.0006$ & $0.0502 \pm 0.0026$
& $0.0524 \pm 0.0016$ & $0.0508 \pm 0.0015$
\\
\end{tabular}
\end{center}
\end{table}
\subsubsection{Evaluation protocol}
\label{app:evaluation}

Generation quality is measured by averaged Maximum Mean Discrepancy (aMMD), the mean MMD over quadratic variation, spectral centroid, and degree correlation. Each MMD sums five RBF kernels whose bandwidths are log-spaced multiples
$10^{-1},\,10^{-1/2},\,1,\,10^{1/2},\,10$ of the median heuristic, following
the released implementation of \citet{rozada2026gad}. Unlike the released GAD evaluation, which uses the symmetric normalized Laplacian for the spectral centroid, our matched protocol uses the combinatorial Laplacian for all graph-based statistics. Each reported cell compares $5{,}000$ generated signals with $5{,}000$ held-out test signals, except for Molene, where both sets contain $150$. All methods use the same normalized signals, graph operators, and metric implementation.
All experiments run on a single Apple M5 Pro (48\,GB unified memory).

\subsubsection{Hyperparameters}
\label{app:hparams}

\begin{table}[ht]
\caption{\textbf{Experimental hyperparameters.}
Values are shared across methods and datasets unless stated otherwise.}
\label{tab:app_hparams}
\begin{center}
\small
\begin{tabular}{@{}lll@{}}
\multicolumn{1}{c}{\bf GROUP} &
\multicolumn{1}{c}{\bf PARAMETER} &
\multicolumn{1}{c}{\bf VALUE}
\\ \hline \\
data     & train / val / test caps & $20{,}000/3{,}000/5{,}000$ \\
     & normalization & per-node z-score, training statistics \\
\hline
graph & symmetrizer & $\tfrac12(\mW+\mW^\top)$ \\
      & self-loops & removed after symmetrization \\
      & Laplacian (model basis) & symmetric normalized \\
      & Laplacian (evaluation)   & combinatorial \\
      & diffusion operator $\mL_\delta$ & $\mL/\lambda_{\max}+\delta\mI$ \\
      & Tikhonov shift $\delta$ & $0.05$ \\
      & backbone graph shift & $\mL/\lambda_{\max}$  \\
\hline
backbone & architecture & \texttt{DenoiserGNN} \citep{rozada2026gad} \\
         & width & $128$ \\
         & blocks $\times$ taps & $3\times5$ \\
         & time embedding & $64$ \\
         & trainable parameters & $61{,}185$, identical for every arm \\
\hline
optimization & optimizer & Adam \\
             & learning rate & $3\times10^{-4}$ \\
             & weight decay & $10^{-4}$ \\
             & batch size & $256$ \\
             & gradient clipping & $1.0$ \\
             & EMA decay & $0.999$ \\
             & validation interval & $250$ updates \\
\hline
stopping & patience & $8$ evaluations \\
         & relative improvement & $0.002$ \\
         & maximum updates & $40{,}000$ \\
\hline
conjugate & $\kappa$, finite SNR & $2.0$ \\
forward   & $\kappa$, near-zero SNR & $32.6356$ (METR-LA), $32.64$ (others) \\
          & $t$ range & $[0.02,1.0]$ \\
          & $\sigma$ & $1.0$ \\
\hline
VE forward & $q_{\min},q_{\max}$ & $0.02,\ 32.64$ \\
           & training draw & log-normal, $P_{\rm mean}{=}-1.2$, $P_{\rm std}{=}1.2$ \\
           & sampling grid exponent $\rho_{\rm VE}$ & $7.0$ \\
\hline
reference & covariance estimator & Ledoit--Wolf \citep{ledoit2004covariance} \\
          & retained covariance & graph-Fourier diagonal $v_i$ \\
          & floor $\varepsilon_{\rm ref}$ & $10^{-6}$ \\
\hline
EDM & $P_{\rm mean},P_{\rm std},\rho$ & $-1.2,\ 1.2,\ 7.0$ \\
    & $\sigma_{\min},\sigma_{\max}$ & $0.002,\ 80.0$ \\
\hline
GAD process & $\gamma,S,\sigma,T$ & $0.8,\ 7.0,\ 1.0,\ 1.0$ \\
            & $\alpha,c_{\min}$ & $4.0,\ 0.1$ \\
            & total-variation loss weight & $5\times10^{-3}$ \\
            & time-sampling exponent & $0.65$ \\
\hline
GAD sampler & Newton iterations & $2$ \\
            & terminal $s$ & $0.0$ \\
\hline
GRCD sampler & residual quadrature & composite Simpson, $65$ nodes \\
             & selected grid exponent $\rho$ & $3$ \\
\hline
evaluation & RBF bandwidth multipliers & $\{0.1,0.32,1,3.2,10\}\times$ median \\
           & generated / test samples & $5{,}000/5{,}000$ ($150/150$ on Molene) \\
           & experimental seeds & $0,1,2$ \\
\end{tabular}
\end{center}
\end{table}

\subsection{Other Benchmark Results}
\label{app:other-benchmark-results}

GRCD is the best method at NFE~4 on all four settings. The margin narrows with budget: by NFE~64 the isotropic and static preconditioning controls and EDM match or beat it on the three SBM settings, as does the conjugate preconditioner on SBM~$c{=}16$, where they converge to a lower floor. On Molene, EDM's NFE-16 entry is a crossing point and it is not a converged value. Its per-node dispersion is still contracting from $45\times$ toward $1$, and the same arm degrades to $0.176$ and $0.213$ at NFE~32 and~64. The three preconditioning controls (Isotropic VE, Static precond.\ and Conjugate precond.)\ share this shape.
In Table~\ref{tab:main} and the tables below, ``GRCD w/o residual score'' uses the fitted terminal reference with \(\epsilon\)-prediction and the Heun sampler. The last two rows add the residual score parameterization and the exponential-residual solver.

\begin{table}[ht]
\caption{\textbf{Matched comparison on Molene.}
aMMD at the validation-converged stage and a fixed nominal NFE budget (Appendix~\ref{app:samplers}). Lower is better and bold denotes
the best mean in each column. Results are mean $\pm$ sample standard
deviation over three random seeds. }
\label{tab:app_molene}
\begin{center}
\small
\setlength{\tabcolsep}{5pt}
\begin{tabular}{@{}lcccc@{}}
\multicolumn{1}{c}{\bf METHOD} &
\multicolumn{1}{c}{\bf NFE 4} &
\multicolumn{1}{c}{\bf NFE 8} &
\multicolumn{1}{c}{\bf NFE 16} &
\multicolumn{1}{c}{\bf NFE 64}
\\ \hline \\
EDM
& $2.6268 \pm 0.0378$ & $1.3517 \pm 0.0869$
& $\mathbf{0.0951 \pm 0.0110}$
& $0.2126 \pm 0.0330$
\\
GAD
& $2.7836 \pm 0.0116$ & $2.7613 \pm 0.0558$
& $2.7147 \pm 0.0155$
& $2.7303 \pm 0.0261$
\\
WSD-style, scalar
& $2.6712 \pm 0.0060$ & $1.5278 \pm 0.0730$
& $1.1318 \pm 0.0384$
& $1.0181 \pm 0.0134$
\\
WSD-style, graph prior
& $2.5191 \pm 0.0101$ & $0.8458 \pm 0.0235$
& $1.4101 \pm 0.0507$
& $1.1717 \pm 0.0493$
\\
Isotropic VE
& $2.6264 \pm 0.0319$ & $0.5675 \pm 0.0779$
& $0.1243 \pm 0.0304$
& $0.2159 \pm 0.0354$
\\
Static precond.
& $2.6326 \pm 0.0287$ & $0.7461 \pm 0.0474$
& $0.1325 \pm 0.0163$
& $0.2208 \pm 0.0209$
\\
Conjugate precond.
& $2.6308 \pm 0.0133$ & $0.6139 \pm 0.0307$
& $0.1377 \pm 0.0130$
& $0.2284 \pm 0.0295$
\\
GRCD w/o residual score
& $0.1545 \pm 0.0295$ & $0.1207 \pm 0.0198$
& $0.1556 \pm 0.0055$
& $0.1716 \pm 0.0078$
\\
GRCD, $\rho{=}1$
& $0.5514 \pm 0.0003$ & $\mathbf{0.1018 \pm 0.0275}$
& $0.1816 \pm 0.0181$
& $\mathbf{0.1599 \pm 0.0245}$
\\
GRCD, $\rho{=}3$
& $\mathbf{0.1114 \pm 0.0497}$ & $0.1731 \pm 0.0308$
& $0.1725 \pm 0.0414$
& $0.1704 \pm 0.0286$
\\
\end{tabular}
\end{center}
\end{table}

\begin{table}[ht]
\caption{\textbf{Matched comparison on SBM ($c{=}1$).}
aMMD at the validation-converged stage and a fixed nominal NFE budget (Appendix~\ref{app:samplers}). Lower is better and bold denotes
the best mean in each column. Results are mean $\pm$ sample standard
deviation over three random seeds. }
\label{tab:app_sbmc1}
\begin{center}
\small
\setlength{\tabcolsep}{5pt}
\begin{tabular}{@{}lcccc@{}}
\multicolumn{1}{c}{\bf METHOD} &
\multicolumn{1}{c}{\bf NFE 4} &
\multicolumn{1}{c}{\bf NFE 8} &
\multicolumn{1}{c}{\bf NFE 16} &
\multicolumn{1}{c}{\bf NFE 64}
\\ \hline \\
EDM
& $1.7509 \pm 0.0082$ & $1.5238 \pm 0.0101$
& $0.4285 \pm 0.0026$
& $0.0119 \pm 0.0017$
\\
GAD
& $2.2376 \pm 0.0018$ & $2.0090 \pm 0.0085$
& $1.9618 \pm 0.0034$
& $1.9395 \pm 0.0078$
\\
WSD-style, scalar
& $1.3046 \pm 0.1772$ & $1.4082 \pm 0.0704$
& $1.3388 \pm 0.0665$
& $0.8715 \pm 0.0535$
\\
WSD-style, graph prior
& $1.2084 \pm 0.2054$ & $1.4998 \pm 0.1438$
& $1.4533 \pm 0.0949$
& $0.8845 \pm 0.0644$
\\
Isotropic VE
& $1.7524 \pm 0.0067$ & $1.2200 \pm 0.0027$
& $0.1704 \pm 0.0014$
& $\mathbf{0.0113 \pm 0.0019}$
\\
Static precond.
& $1.7522 \pm 0.0083$ & $1.2339 \pm 0.0152$
& $0.1465 \pm 0.0255$
& $0.0210 \pm 0.0073$
\\
Conjugate precond.
& $1.7521 \pm 0.0062$ & $1.1663 \pm 0.0203$
& $0.1835 \pm 0.0158$
& $0.0621 \pm 0.0455$
\\
GRCD w/o residual score
& $0.0527 \pm 0.0092$ & $\mathbf{0.0426 \pm 0.0050}$
& $\mathbf{0.0427 \pm 0.0050}$
& $0.0429 \pm 0.0051$
\\
GRCD, $\rho{=}1$
& $\mathbf{0.0485 \pm 0.0124}$ & $0.0502 \pm 0.0095$
& $0.0508 \pm 0.0086$
& $0.0510 \pm 0.0083$
\\
GRCD, $\rho{=}3$
& $0.0497 \pm 0.0062$ & $0.0494 \pm 0.0029$
& $0.0526 \pm 0.0069$
& $0.0502 \pm 0.0030$
\\
\end{tabular}
\end{center}
\end{table}

\begin{table}[ht]
\caption{\textbf{Matched comparison on SBM ($c{=}4$).}
aMMD at the validation-converged stage and a fixed nominal NFE budget (Appendix~\ref{app:samplers}). Lower is better and bold denotes
the best mean in each column. Results are mean $\pm$ sample standard
deviation over three random seeds. }
\label{tab:app_sbmc4}
\begin{center}
\small
\setlength{\tabcolsep}{5pt}
\begin{tabular}{@{}lcccc@{}}
\multicolumn{1}{c}{\bf METHOD} &
\multicolumn{1}{c}{\bf NFE 4} &
\multicolumn{1}{c}{\bf NFE 8} &
\multicolumn{1}{c}{\bf NFE 16} &
\multicolumn{1}{c}{\bf NFE 64}
\\ \hline \\
EDM
& $1.9471 \pm 0.0082$ & $1.6387 \pm 0.0103$
& $0.4467 \pm 0.0045$
& $0.0161 \pm 0.0028$
\\
GAD
& $2.3262 \pm 0.0017$ & $2.1619 \pm 0.0091$
& $2.1281 \pm 0.0028$
& $2.1100 \pm 0.0066$
\\
WSD-style, scalar
& $1.5867 \pm 0.2069$ & $1.3181 \pm 0.1055$
& $1.4083 \pm 0.1201$
& $0.9278 \pm 0.0792$
\\
WSD-style, graph prior
& $1.4519 \pm 0.2599$ & $1.3647 \pm 0.2030$
& $1.5479 \pm 0.1677$
& $0.9285 \pm 0.1027$
\\
Isotropic VE
& $1.9492 \pm 0.0062$ & $1.2787 \pm 0.0074$
& $0.1797 \pm 0.0081$
& $0.0164 \pm 0.0041$
\\
Static precond.
& $1.9518 \pm 0.0061$ & $1.3265 \pm 0.0086$
& $0.1766 \pm 0.0115$
& $\mathbf{0.0132 \pm 0.0035}$
\\
Conjugate precond.
& $1.9491 \pm 0.0073$ & $1.1984 \pm 0.0346$
& $0.1927 \pm 0.0115$
& $0.0696 \pm 0.0487$
\\
GRCD w/o residual score
& $0.0707 \pm 0.0203$ & $\mathbf{0.0525 \pm 0.0105}$
& $\mathbf{0.0521 \pm 0.0100}$
& $0.0522 \pm 0.0101$
\\
GRCD, $\rho{=}1$
& $\mathbf{0.0574 \pm 0.0141}$ & $0.0641 \pm 0.0128$
& $0.0665 \pm 0.0118$
& $0.0673 \pm 0.0114$
\\
GRCD, $\rho{=}3$
& $0.0616 \pm 0.0080$ & $0.0646 \pm 0.0037$
& $0.0685 \pm 0.0092$
& $0.0655 \pm 0.0046$
\\
\end{tabular}
\end{center}
\end{table}

\begin{table}[ht]
\caption{\textbf{Matched comparison on SBM ($c{=}16$).}
aMMD at the validation-converged stage and a fixed nominal NFE budget (Appendix~\ref{app:samplers}). Lower is better and bold denotes
the best mean in each column. Results are mean $\pm$ sample standard
deviation over three random seeds. }
\label{tab:app_sbmc16}
\begin{center}
\small
\setlength{\tabcolsep}{5pt}
\begin{tabular}{@{}lcccc@{}}
\multicolumn{1}{c}{\bf METHOD} &
\multicolumn{1}{c}{\bf NFE 4} &
\multicolumn{1}{c}{\bf NFE 8} &
\multicolumn{1}{c}{\bf NFE 16} &
\multicolumn{1}{c}{\bf NFE 64}
\\ \hline \\
EDM
& $2.1231 \pm 0.0086$ & $1.7378 \pm 0.0141$
& $0.4586 \pm 0.0183$
& $0.0204 \pm 0.0049$
\\
GAD
& $2.4178 \pm 0.0017$ & $2.2966 \pm 0.0091$
& $2.2726 \pm 0.0023$
& $2.2574 \pm 0.0064$
\\
WSD-style, scalar
& $1.5899 \pm 0.4909$ & $1.6443 \pm 0.5671$
& $1.4122 \pm 0.2295$
& $0.8368 \pm 0.2536$
\\
WSD-style, graph prior
& $1.5082 \pm 0.4521$ & $1.7000 \pm 0.7449$
& $1.6301 \pm 0.2184$
& $0.8894 \pm 0.1747$
\\
Isotropic VE
& $2.1257 \pm 0.0056$ & $1.3307 \pm 0.0081$
& $0.1824 \pm 0.0181$
& $0.0202 \pm 0.0054$
\\
Static precond.
& $2.1269 \pm 0.0058$ & $1.3865 \pm 0.0251$
& $0.1758 \pm 0.0201$
& $0.0136 \pm 0.0072$
\\
Conjugate precond.
& $2.1273 \pm 0.0064$ & $1.2838 \pm 0.0019$
& $0.1665 \pm 0.0239$
& $\mathbf{0.0093 \pm 0.0072}$
\\
GRCD w/o residual score
& $0.0918 \pm 0.0298$ & $\mathbf{0.0503 \pm 0.0210}$
& $\mathbf{0.0500 \pm 0.0196}$
& $0.0500 \pm 0.0195$
\\
GRCD, $\rho{=}1$
& $\mathbf{0.0270 \pm 0.0121}$ & $0.0530 \pm 0.0089$
& $0.0630 \pm 0.0081$
& $0.0663 \pm 0.0078$
\\
GRCD, $\rho{=}3$
& $0.0421 \pm 0.0144$ & $0.0595 \pm 0.0113$
& $0.0657 \pm 0.0071$
& $0.0638 \pm 0.0096$
\\
\end{tabular}
\end{center}
\end{table}

\subsection{Other Results}
\label{app:other-results}

\subsubsection{Non-learned Gaussian controls}
\label{app:gaussian_control}

The three evaluation statistics are strongly influenced by second-order structure, so a Gaussian with a well-estimated covariance could potentially achieve a low aMMD without learning the non-Gaussian data distribution. We test this directly. The interpretation was fixed before the numbers existed: a control landing near the trained models would mean these three statistics cannot separate covariance matching from learned generation.

We evaluate four zero-mean Gaussian controls, sampled directly without a network, reverse process, or sampling budget: a per-node diagonal Gaussian $\mathcal{N}(\vzero,\operatorname{diag}(\mSigma_{\mathrm{LW}}))$, the full Ledoit--Wolf Gaussian $\mathcal{N}(\vzero,\mSigma_{\mathrm{LW}})$, the GMRF $\mathcal{N}(\vzero,\sigma^2\mL_\delta^{-1})$ corresponding to the near-zero-SNR limit in \Eqref{eq:gmrf_limit_main}, and $\mathcal{N}(\vzero,\mU\operatorname{diag}(\vv)\mU^\top)$, the fitted graph-spectral Gaussian reference before forward propagation. Their samples are evaluated with the same aMMD estimator and held-out test sets as the trained models. Because these controls have no training randomness, their reported spread is computed over $5$ independent draws.

Table~\ref{tab:app_gaussian} shows that covariance matching alone does not account for GRCD's low-NFE performance: on every setting, the best Gaussian control has higher aMMD than GRCD at NFE~$4$. On METR-LA the strongest control reaches $0.1921$ against GRCD's $0.0538$, and the graph-spectral reference alone reaches $0.4134$. Within the three SBM settings, the gap increases with spectral concentration, from $1.08\times$ at $c=1$ to $2.93\times$ at $c=16$.

On the SBM settings, GRCD reaches a plateau near its low aMMD values by moderate NFE, while alternative numerical solvers plateau at similar levels, suggesting that this behavior is not primarily caused by the numerical integrator. At $c=1$, the fitted graph-spectral Gaussian reference itself reaches $0.0535$, close to GRCD's $0.0497$, indicating that the learned residual provides only a small additional improvement under this metric in that regime. In contrast, EDM and the isotropic and static preconditioning controls continue improving at larger budgets and eventually outperform GRCD at NFE~$64$.

\begin{table}[ht]
\caption{Gaussian controls versus GRCD at NFE~$4$. aMMD. Lower is better. Gaussian-control values are mean $\pm$ sample standard deviation over $5$ independent draws. ``Best control'' denotes the lowest aMMD among the four non-learned Gaussian controls. In the brackets, ``full'' is the full Ledoit--Wolf Gaussian and ``spectral'' is the fitted graph-spectral Gaussian
reference.}
\label{tab:app_gaussian}
\begin{center}
\small
\setlength{\tabcolsep}{6pt}
\begin{tabular}{@{}lccc@{}}
\multicolumn{1}{c}{\bf SETTING} &
\multicolumn{1}{c}{\bf BEST CONTROL} &
\multicolumn{1}{c}{\bf GRCD} &
\multicolumn{1}{c}{\bf RATIO}
\\ \hline \\
METR-LA        & $0.1921 \pm 0.0014$ {\scriptsize(full)}     & $\mathbf{0.0538}$ & $\mathbf{3.57\times}$ \\
SBM ($c{=}16$) & $0.1232 \pm 0.0041$ {\scriptsize(spectral)}    & $\mathbf{0.0421}$ & $\mathbf{2.93\times}$ \\
Molene         & $0.1822 \pm 0.0242$ {\scriptsize(spectral)}    & $\mathbf{0.1114}$ & $\mathbf{1.64\times}$ \\
SBM ($c{=}4$)  & $0.0799 \pm 0.0033$ {\scriptsize(spectral)}    & $\mathbf{0.0616}$ & $\mathbf{1.30\times}$ \\
SBM ($c{=}1$)  & $0.0535 \pm 0.0027$ {\scriptsize(spectral)}    & $\mathbf{0.0497}$ & $\mathbf{1.08\times}$ \\
\end{tabular}
\end{center}
\end{table}

\subsubsection{Cost to reach a quality target}
\label{app:cost_target}
\begin{table}[t]
\caption{\textbf{Cost to reach a quality target on METR-LA.} Cheapest measured
sampling cost, in seconds per $1000$ samples, at which each configuration first reaches an aMMD target at $\kappa=2$, converged. ``never'' means the target is not reached at any budget up to NFE~$64$. ``$+$ fitted terminal reference'' is the row
called ``GRCD w/o residual score'' in Table~\ref{tab:main}.}
\label{tab:cost_target}
\begin{center}
\small
\setlength{\tabcolsep}{5pt}
\begin{tabular}{@{}lccc@{}}
\multicolumn{1}{c}{\bf CONFIGURATION} &
\multicolumn{1}{c}{\bf aMMD $\le 0.2$} &
\multicolumn{1}{c}{\bf aMMD $\le 0.1$} &
\multicolumn{1}{c}{\bf aMMD $\le 0.06$}
\\ \hline \\
EDM                        & $4.01$ {\scriptsize(NFE 32)} & never & never \\
GAD                        & never & never & never \\
WSD-style, scalar          & never & never & never \\
WSD-style, graph prior     & $2.07$ {\scriptsize(NFE 16)} & $8.27$ {\scriptsize(NFE 64)} & never \\
Isotropic VE               & $1.94$ {\scriptsize(NFE 16)} & never & never \\
Static precond.            & $1.94$ {\scriptsize(NFE 16)} & never & never \\
Conjugate precond.         & $1.95$ {\scriptsize(NFE 16)} & $4.02$ {\scriptsize(NFE 32)} & never \\
Mode-dependent clock             & never & never & never \\
\quad + fitted terminal reference & $\mathbf{0.52}$ {\scriptsize(NFE 4)} & $0.78$ {\scriptsize(NFE 6)} & $1.56$ {\scriptsize(NFE 12)} \\
\quad + residual score     & $\mathbf{0.52}$ {\scriptsize(NFE 4)} & $0.78$ {\scriptsize(NFE 6)} & $1.56$ {\scriptsize(NFE 12)} \\
GRCD, $\rho{=}1$   & $0.78$ {\scriptsize(NFE 6)} & $0.78$ {\scriptsize(NFE 6)} & $1.04$ {\scriptsize(NFE 8)} \\
GRCD, $\rho{=}3$ (ours)   & $\mathbf{0.52}$ {\scriptsize(NFE 4)} & $\mathbf{0.52}$ {\scriptsize(NFE 4)} & $\mathbf{0.52}$ {\scriptsize(NFE 4)} \\
\end{tabular}
\end{center}
\end{table}

\end{document}